\documentclass[10pt,twocolumn,letterpaper]{article}

\usepackage[pagenumbers]{iccv} 

\usepackage{comment}
\usepackage{multirow}

\newcommand{\dq}[1]{``#1''}

\definecolor{Set1-Blue}{RGB}{55,126,184}
\definecolor{Set1-Green}{RGB}{77,175,74}
\definecolor{Set1-Orange}{RGB}{255,127,0}

\makeatletter
\newcommand{\DisableBackref}{\let\Hy@backout\@gobble}
\newcommand{\EnableBackref}{\let\Hy@backout\@firstofone}
\makeatother

\definecolor{iccvblue}{rgb}{0.21,0.49,0.74}
\usepackage[pagebackref,breaklinks,colorlinks,allcolors=iccvblue]{hyperref}

\usepackage[accsupp]{axessibility}  

\def\paperID{11577} 
\def\confName{ICCV}
\def\confYear{2025}

\title{DACoN: DINO for Anime Paint Bucket Colorization \\ with Any Number of Reference Images}

\author{Kazuma Nagata \qquad Naoshi Kaneko\\
Tokyo Denki University\\
{\tt\small {24fmi19@ms.dendai.ac.jp, naoshi.kaneko@mail.dendai.ac.jp}}
}

\begin{document}

\twocolumn[{%
\renewcommand\twocolumn[1][]{#1}%
\maketitle
\vspace{-3em} 
\begin{center}
    \centering
    \captionsetup{type=figure}
    \includegraphics[width=1.0\linewidth]{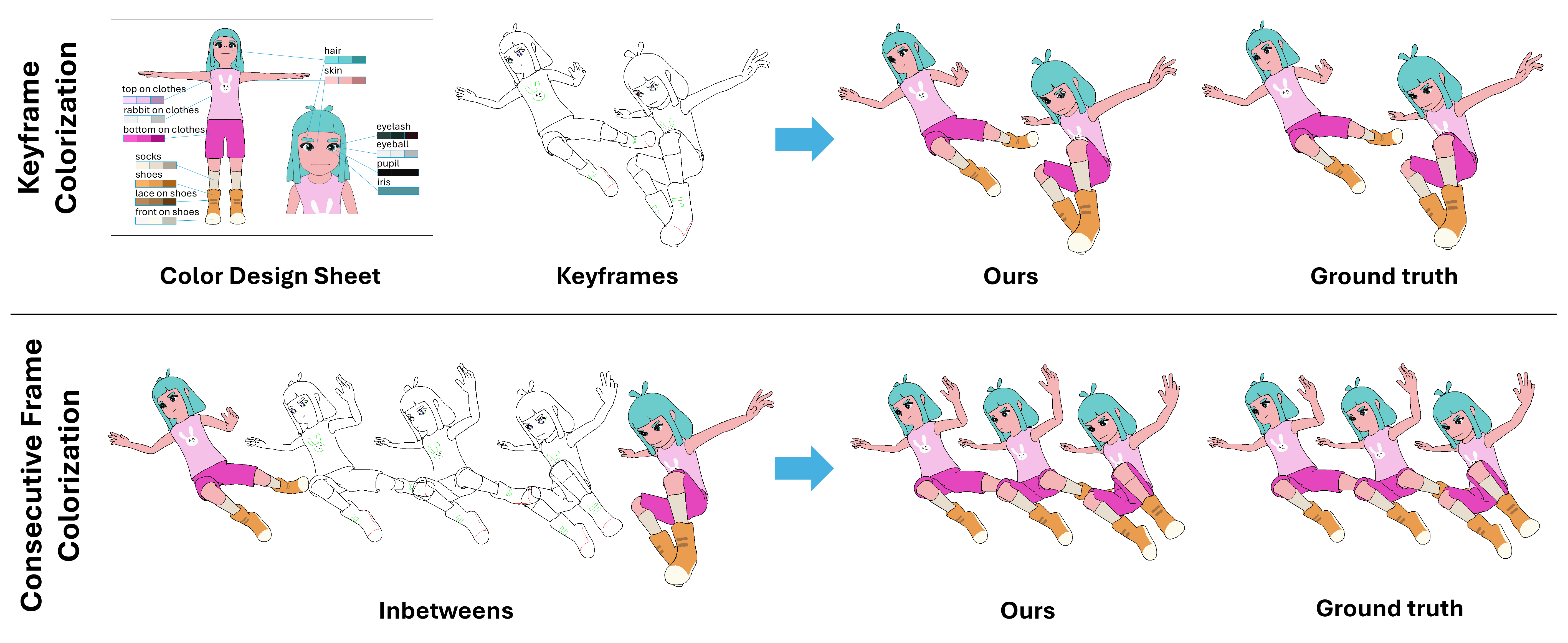}
    \captionof{figure}{DACoN can colorize both \textit{keyframes}, even when their compositions differ significantly from the reference images, and \textit{inbetweens} that interpolate between keyframes, all with a single model.}
    \label{fig:paper_overview}
\end{center}%
}]

\begin{abstract}
Automatic colorization of line drawings has been widely studied to reduce the labor cost of hand-drawn anime production. 
Deep learning approaches, including image/video generation and feature-based correspondence, have improved accuracy but struggle with occlusions, pose variations, and viewpoint changes. 
To address these challenges, we propose DACoN, a framework that leverages foundation models to capture part-level semantics, even in line drawings. 
Our method fuses low-resolution semantic features from foundation models with high-resolution spatial features from CNNs for fine-grained yet robust feature extraction.
In contrast to previous methods that rely on the Multiplex Transformer and support only one or two reference images, DACoN removes this constraint, allowing any number of references. 
Quantitative and qualitative evaluations demonstrate the benefits of using multiple reference images, achieving superior colorization performance.
Our code and model are available at \href{https://github.com/kzmngt/DACoN}{https://github.com/kzmngt/DACoN}.
\end{abstract}  
\section{Introduction}
\label{sec:intro}
In anime production, animators color line drawings as part of the workflow.
As shown in Figure~\ref{fig:paper_overview}, anime line drawings consist of keyframes, which capture critical moments in motion, and inbetweens, which interpolate frames between these keyframes~\cite{tang2025generative}.
Keyframes are manually colorized using color design sheets, which provide a standardized reference for character colors, while inbetweens rely on the colorized keyframes.
These tasks are referred to as \textit{keyframe colorization} and \textit{consecutive frame colorization}, respectively~\cite{dai2024paint}.

The core manual process for anime colorization is often termed \textit{paint bucket colorization}, which specifically involves painters using a paint bucket tool and a designated color palette to fill completely line-enclosed segments. 
While various digital drawing tools have been developed to improve efficiency, color assignment still relies heavily on manual work. 
Automatic colorization methods have been studied extensively and can be broadly categorized into pixel-based~\cite{lee2020reference,wu2023self,lin2024visual,liu2025manganinja,zhuang2024colorflow,xing2024tooncrafter,huang2024lvcd,yang2025layeranimate,meng2024anidoc} and segment-based approaches~\cite{dai2024paint,shaolong2023shape,chen2020active,maejima2019graph,liu2020shape,zhu2016globally,casey2021animation,dai2024learning}. 
The former employs techniques such as color transfer or generative models, while the latter relies on graph optimization or feature matching.

Previous studies primarily focus on consecutive frame colorization, leveraging spatial similarity between frames to propagate colors~\cite{casey2021animation,dai2024learning,dang2020correspondence}. However, these methods struggle when a reference image lacks the necessary parts for color inference. Human animators overcome this issue by referring to color design sheets or different colorized frames, suggesting the importance of utilizing multiple references.
Nevertheless, existing approaches are typically limited to one or two reference images due to the constraints of the Multiplex Transformer~\cite{sarlin2020superglue}, a mechanism that enhances colorization by aggregating feature representations from multiple inputs via Cross-Attention. While this design allows for effective feature fusion, it also imposes a fixed input structure, restricting the number of reference images that can be incorporated. This limitation hinders the ability to infer missing colors from diverse references, making automatic colorization less flexible in real-world scenarios.

In contrast, keyframe colorization remains underexplored, primarily because large compositional differences (\eg, changes in pose, viewpoint, and occlusion) between reference and target images hinder spatial alignment. Unlike natural images, line drawings consist only of contours, often stylized or deformed, making it difficult to extract meaningful semantic features. Consequently, the accuracy of keyframe colorization remains lower than that of consecutive frame colorization~\cite{dai2024paint}.

\begin{figure}[t]
    \centering
    \includegraphics[width=1.0\linewidth]{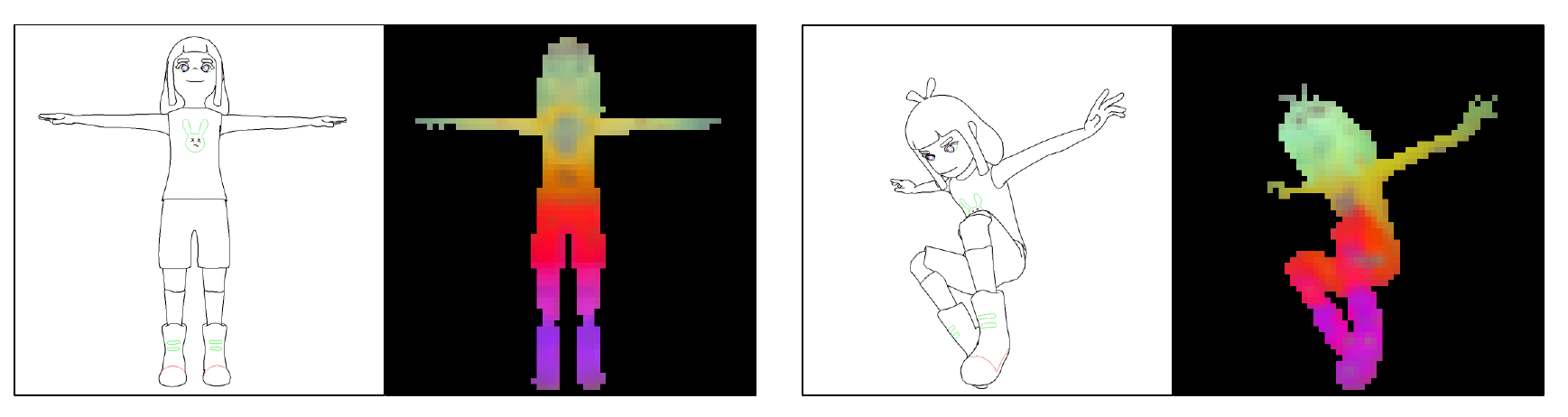}
    \caption{Visualization of DINO features. The figure shows two pairs of images. In each pair, the left image is the input, and the right image is the PCA visualization of DINO features after background removal.}
    \vspace{-0.2cm}
    \label{fig:dinov2_pca_visualization}
\end{figure}

To address these challenges, we focus on the feature representations of foundation models. As shown in Figure~\ref{fig:dinov2_pca_visualization}, DINOv2~\cite{oquab2023dinov2} captures part-level semantic information not only for natural images but also for line drawings. These properties suggest that DINO features can serve as an effective guide for automatic colorization through correspondence. However, since DINOv2 extracts features in low-resolution patches, it struggles to capture fine-grained details.
To enhance local feature representations, we propose DACoN, a network that integrates DINOv2 with U-Net~\cite{ronneberger2015u} to refine spatial features. Instead of relying on conventional feature aggregation, we design a framework that prioritizes utilizing diverse reference information, inspired by the zero-shot correspondence capability of foundation models.
Additionally, we introduce a Feature Consistency Loss to preserve the integrity of DINO features, further improving colorization accuracy.
Extensive experiments show that DACoN achieves higher accuracy than existing methods in both keyframe and consecutive frame colorization with a unified model and training, notably improving keyframe colorization.
Our contributions in this study can be summarized as follows:

\begin{itemize}
    \item We introduce the foundation model DINOv2 for anime paint bucket colorization, achieving superior accuracy compared to previous methods.
    \item We propose a network combining DINOv2 and U-Net, capable of handling both keyframe and consecutive frame colorization.
    \item We facilitate correspondence with multiple reference images, demonstrating that incorporating additional references enhances colorization performance. 
\end{itemize}

\begin{figure*}[t]
    \centering
    \includegraphics[width=1.0\linewidth]{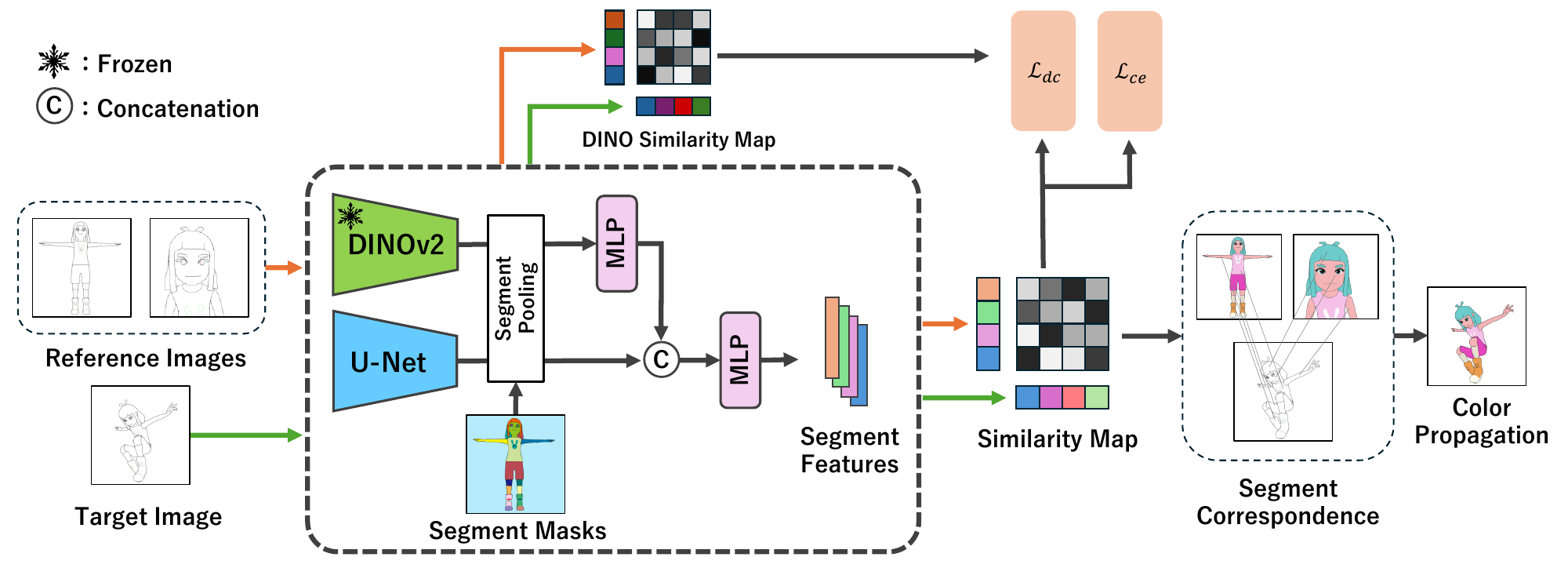}
    \caption{Overview of DACoN pipeline. Each image is individually input into the model to extract segment features. Since the model does not reference other images, there is no limit on the number of images used for segment correspondence.}
    \label{fig:architecture}
\end{figure*}

\section{Related Work}
\label{sec:related}

\subsection{Color Propagation via Segment Correspondence}

Segment correspondence-based methods transfer colors from a reference to a target line image by establishing segment-wise correspondences, categorized into graph-based and feature matching-based approaches.

Graph-based methods represent segments as graph nodes and establish correspondences while preserving shape and spatial consistency~\cite{shaolong2023shape,chen2020active,maejima2019graph,liu2020shape,zhu2016globally,sato2014reference}. 
Liu~\etal~\cite{shaolong2023shape} proposed a method leveraging Kendall shape space for shape and positional similarity, optimized via spectral matching. 
However, these methods suffer from high computational costs as the number of segments increases.

Feature matching-based methods extract segment-wise features and propagate colors via nearest-neighbor correspondences~\cite{dai2024paint,casey2021animation,dai2024learning,dang2020correspondence}. 
AnT~\cite{casey2021animation} enhances segment-wise feature representation by refining CNN features through a Multiplex Transformer~\cite{sarlin2020superglue}.
BasicPBC~\cite{dai2024paint,dai2024learning} employs optical flow for temporal consistency in consecutive frames, while utilizing text descriptions and color design sheets for keyframe colorization.
Nevertheless, these methods depend on a single reference image, failing when the required segment is absent.

\subsection{Colorization via Semantic Segmentation}

Semantic segmentation-based methods classify each pixel in a line drawing into target color classes~\cite{ramassamy2018pre,maejima2024continual}.
Ramassamy~\etal~\cite{ramassamy2018pre} used an FCN for pixel-level color prediction, refining results by majority voting within segments.
Maejima~\etal~\cite{maejima2024continual} introduced an efficient patch-based approach, restricting sampling to line art corners.
However, these methods are character-specific, requiring separate models for each character.

\subsection{Colorization via Image/Video Generation}

Image~\cite{lin2024visual,liu2025manganinja,zhuang2024colorflow,cao2024animediffusion,furusawa2017comicolorization,shi2022reference,yifeng2024animation,ruizhi2021line,li2022eliminating} and video~\cite{xing2024tooncrafter,huang2024lvcd,yang2025layeranimate,meng2024anidoc,shi2020deep,loftsdottir2022sketchbetween,wang2023coloring,zhang2021line} generation-based methods leverage reference color images and line drawings to condition a generative model for colorization.

MangaNinja~\cite{liu2025manganinja} employs a diffusion-based model with patch shuffling and point-driven control, allowing user-specified correspondences. 
Similarly, ColorFlow~\cite{zhuang2024colorflow} partitions the target into regions, retrieves similar references, and merges them into a single input for colorization and super-resolution.
For video generation, 
AniDoc~\cite{meng2024anidoc} preserves color consistency under viewpoint and layout changes by conditioning the generative process on keypoint trajectories.
LVCD~\cite{huang2024lvcd} leverages previously colorized frames and future line drawings with blending and attention mechanisms to maintain temporal consistency.

Although these methods can handle significant compositional variations, they may generate colors not present in the reference images.
Furthermore, generating composite images or videos instead of segmented layers makes post-processing and fine-grained adjustments difficult, limiting their use in anime production.

\subsection{Animation-Related Datasets}
So far, various anime-related datasets have been introduced, covering both hand-drawn~\cite{siyao2021deep,dai2024learning} and CG-rendered~\cite{shugrina2019creative,siyao2022animerun,dai2024learning} data. 
Creative Flow+~\cite{shugrina2019creative} provides rendered images with annotations such as optical flow, correspondences, and object IDs. 
AnimeRun~\cite{siyao2022animerun} converts open-source 3D animations into a 2D anime style, offering per-pixel optical flow and segment-level IDs. 
However, anti-aliasing in both datasets hinders precise per-segment color extraction. Moreover, AnimeRun relies on rendered optical flow, which may result in multiple plausible ground truth assignments for a single color segment. This ambiguity reduces benchmark reliability.
Dai~\etal~\cite{dai2024learning, dai2024paint} introduced the PaintBucket-Character dataset, a large and diverse corpus tailored for anime paint bucket colorization. 
It includes both rendered and hand-drawn data and contains only characters without backgrounds, a standard practice in anime production. 
It also provides annotations for shading and highlight regions, along with color design sheets featuring images from various angles and distances. 
We use the PaintBucket-Character dataset for training and evaluation.
\section{Method}
\label{sec:method}
Figure~\ref{fig:architecture} provides an overview of our method. 
Given $K$ reference images and a target image, both consisting of line drawings, let the reference images be denoted by $L_k \in \mathbb{R}^{H \times W \times 3}$ for $k=1,...,K$, and the target image by $L_t \in \mathbb{R}^{H \times W \times 3}$.
Their corresponding segment masks, which represent regions enclosed by continuous line boundaries, are given by $m_k \in \mathbb{R}^{M_k \times H \times W}$ for each reference image and $m_t\in \mathbb{R}^{N \times H \times W}$ for the target image.
Each reference image is also associated with its segment color information $c_k \in \mathbb{R}^{M_k \times 3}$.
The total number of segments across all reference images is defined as $M = \sum_{k=1}^{K} M_k$.
We extract features from these $M$ segments in the reference images and $N$ segments in the target image.

\subsection{Architecture}

\noindent{\textbf{Feature Extraction.}}
We utilize the DINOv2 encoder~\cite{oquab2023dinov2} to obtain semantic information, and a CNN-based U-Net~\cite{ronneberger2015u} to extract pixel-level spatial details.
The DINOv2 encoder is frozen during training.
For both reference and target images, the line drawings $L_k$ (for $k=1,...,K$) and $L_t$ are fed into both the DINOv2 encoder and the U-Net.
The DINOv2 encoder produces $C_d$-dimensional DINO feature maps 
$D_k \in \mathbb{R}^{H_d \times W_d \times C_d}$ for each reference image and 
$D_t \in \mathbb{R}^{H_d \times W_d \times C_d}$ for the target image.
Similarly, the U-Net generates $C_u$-dimensional CNN feature maps 
$U_k, U_t \in \mathbb{R}^{H \times W \times C_u}$.
In our experiments, input images are resized to $518 \times 518$ for the DINOv2 encoder and $512 \times 512$ for the U-Net.
We set $C_d = 1024$ and $C_u = 128$.

\noindent{\textbf{Segment Pooling.}}
Following the concept of superpixel pooling~\cite{schuurmans2018efficient}, we extract segment-wise features. Specifically, we first resize $D_k$, $D_t$, $U_k$, and $U_t$ to the original image size via bilinear interpolation. 
Then, segment-wise DINO features 
$d_k \in \mathbb{R}^{M_k \times C_d}$ (for reference images) and 
$d_t \in \mathbb{R}^{N \times C_d}$ (for the target image),
and CNN features 
$u_k \in \mathbb{R}^{M_k \times C_u}$ and
$u_t \in \mathbb{R}^{N \times C_u}$
are computed.
We first apply the segment mask to the corresponding feature maps using element-wise multiplication, and then compute the average over the spatial dimensions.
Formally, for each segment, we compute
\begin{equation}
  \begin{aligned}
    d &= \operatorname{avg}(D \odot m), \\
    u &= \operatorname{avg}(U \odot m).
  \end{aligned}
\end{equation}
where $D, U$ denote either $D_k, U_k$ for reference images or $D_t, U_t$ for the target image, and $m$ represents the corresponding segment masks $m_k$ or $m_t$.

\noindent{\textbf{Fusion of DINO Features and CNN Features.}}
The DINO and CNN features are fused into a unified representation.
First, an MLP reduces the dimensionality of the DINO features $d_k$ and $d_t$ from $C_d$ to $C_u$. 
Next, the reduced DINO features are concatenated with the corresponding CNN features, followed by another MLP to obtain the final feature representations $f_k \in \mathbb{R}^{M_k \times C_u}$ and $f_t \in \mathbb{R}^{N \times C_u}$. 

\noindent{\textbf{Segment Correspondence and Color Propagation.}}  
We first concatenate $f_k \in \mathbb{R}^{M_k \times C_u}$ across all reference images along the segment dimension, forming $f_r \in \mathbb{R}^{M \times C_u}$, where $M = \sum_{k=1}^{K} M_k$.
Similarly, we concatenate the corresponding segment color information $c_k \in \mathbb{R}^{M_k \times 3}$, resulting in $c_r \in \mathbb{R}^{M \times 3}$. 
The cosine similarity between the reference features $f_r$ and the target segment features $f_t$ is then computed to form a similarity map $\hat{S} \in \mathbb{R}^{M \times N}$.
For each target segment, the most similar reference segment is identified using $\operatorname{argmax}$ over the similarity scores, and the corresponding color values from $c_r$ are propagated to predict the target segment colors $\hat{c}_t \in \mathbb{R}^{N \times 3}$.

\subsection{Loss}

\noindent{\textbf{Cross-Entropy Loss.}}
To frame the colorization task as a classification problem, predicted colors are treated as discrete classes. The proposed model is trained using reference images to learn accurate colorization.
First, we apply a temperature-scaled softmax function with $T=0.1$ to the segment feature similarity map, enhancing the contrast of the logits and producing a sharper probability distribution over reference image segments.
Then, the probabilities of segments with the same color are summed to derive the color probability distribution $\hat{p}$. 
Finally, the cross-entropy loss $\mathcal{L}_{\mathrm{ce}}$ is computed based on the one-hot representation of the ground truth colors $p$.
Regions where the ground truth color is absent in the reference image are excluded from the loss computation.
Here, $C$ denotes the number of colors.
\begin{align}
\mathcal{L}_{\text{ce}} &= - \sum_{n=1}^{N} \sum_{c=1}^{C} p_{n,c} \log \hat{p}_{n,c}.
\end{align}

\noindent{\textbf{DINO-guided Feature Consistency Loss.}}
DINOv2 captures global features, enabling semantically similar regions to exhibit similar representations, even when their colors (labels) differ. 
To preserve these features, the DINO-guided Feature Consistency Loss $\mathcal{L}_{\mathrm{dc}}$ is introduced.
This loss encourages consistency between the feature similarities computed from DINOv2 and those derived from our proposed model.

Following the same procedure used to obtain the final similarity map $\hat{S}$, we compute the cosine similarity between the DINO features of each segment in the reference and target images, denoted as $d_r$ and $d_t$, to obtain the similarity map $\hat{S}^{\prime} \in \mathbb{R}^{M \times N}$. The final feature similarity map $\hat{S}$ from the proposed model is then compared with $\hat{S}^{\prime}$, both scaled by a factor of 10 to adjust the magnitude of the loss (since the raw loss values tend to be very small).
The loss is computed using the mean absolute error as:
\begin{align}
\mathcal{L}_{\mathrm{dc}} = \frac{1}{MN} \sum_{m=1}^{M} \sum_{n=1}^{N} \left| \hat{S}_{m,n} - \hat{S}'_{m,n} \right|.
\end{align}

Finally, the overall loss function is formed as a weighted sum of the two losses:
\begin{align}
\mathcal{L}_{\text{final}} = \lambda_{\mathrm{ce}} \mathcal{L}_{\mathrm{ce}} + \lambda_{\mathrm{dc}} \mathcal{L}_{\mathrm{dc}},
\end{align}
where $\lambda_{\mathrm{ce}}$ and $\lambda_{\mathrm{dc}}$ are hyperparameters controlling the contributions of $\mathcal{L}_{\mathrm{ce}}$ and $\mathcal{L}_{\mathrm{dc}}$, respectively. 
In our experiments, we set $\lambda_{\mathrm{ce}} = 0.5$ and $\lambda_{\mathrm{dc}} = 0.2$.

\begin{figure*}[t]
    \centering
    \includegraphics[width=1.0\linewidth]{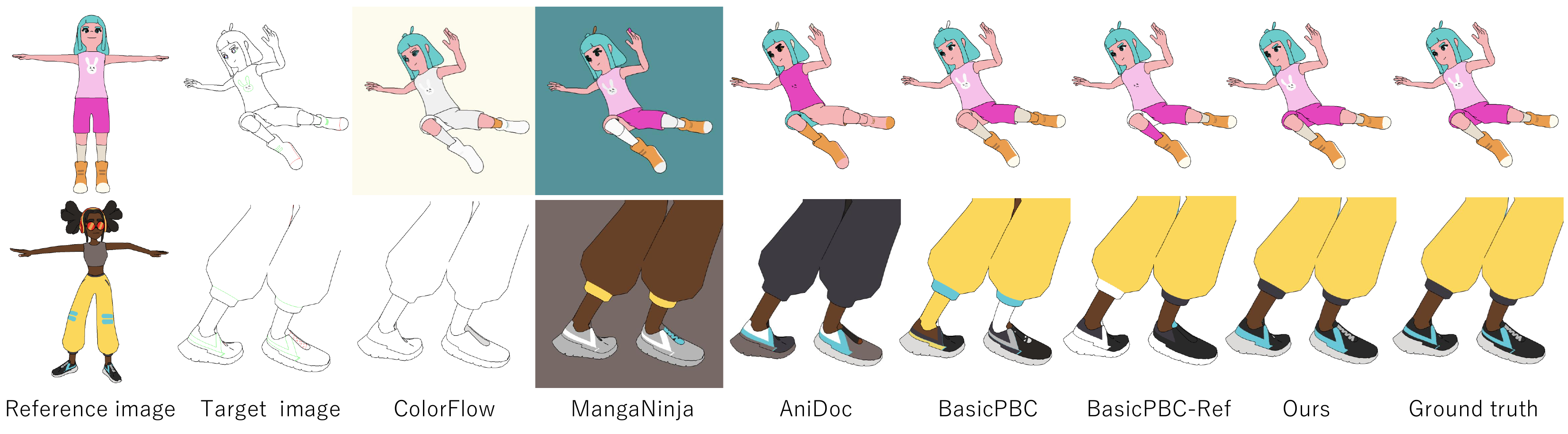}
    \caption{Visual comparisons of keyframe colorization results between our method and other methods.}
    \label{fig:keyframe_compare}
\end{figure*}

\begin{figure}[t]
    \centering
    \includegraphics[width=1.0\linewidth]{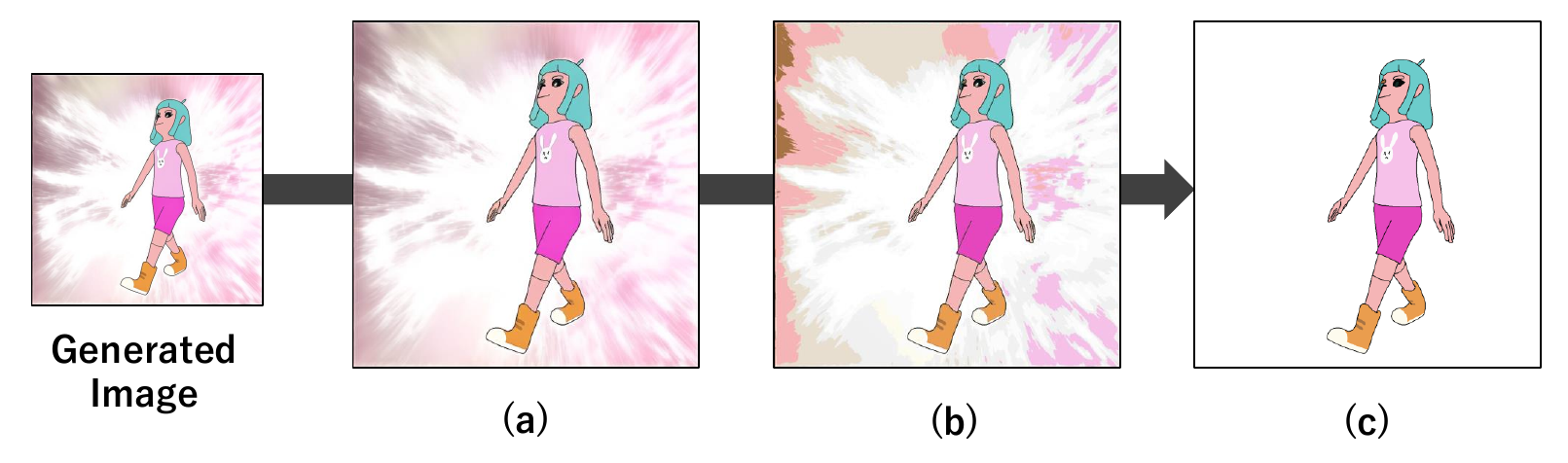}
    \caption{Post-processing of colorization results for generation-based methods. (a) Resize to the original image size. (b) Replacing each pixel with the nearest color from the reference image. (c) Unifying the color to the most frequent one within each segment.}
    \vspace{-0.2cm}
    \label{fig:generate_method_post_process}
\end{figure}

\section{Experiments}
\label{sec:experiments}

\begin{figure*}[t]
    \centering
    \includegraphics[width=1.0\linewidth]{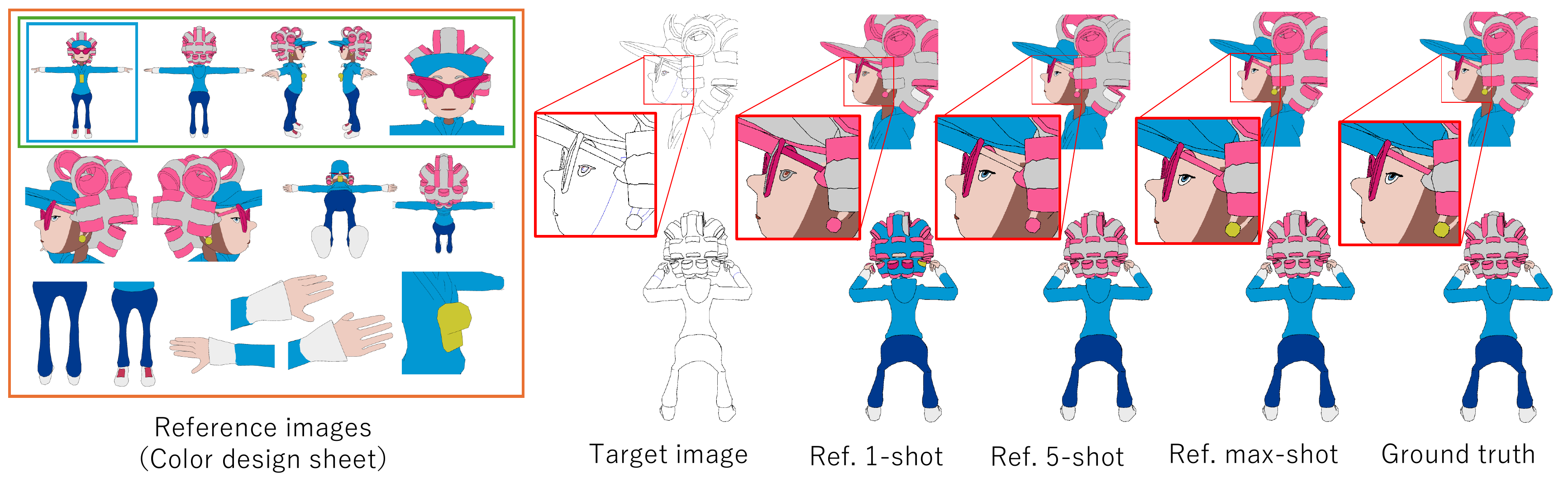}
    \caption{Visual comparison of keyframe colorization results with different numbers of reference images. Images with one, five, and all reference images are framed in blue, green, and orange, respectively.}
    \label{fig:keyframe_reference_counts}
\end{figure*}

\subsection{Dataset}
We use the PaintBucket-Character Dataset released by BasicPBC~\cite{dai2024paint,dai2024learning}. 
This dataset comprises both 3D-rendered and hand-drawn data. 
The 3D-rendered set includes 22 character models performing five actions, captured from three camera angles across 15 video clips. 
Additionally, 9 to 16 images per character are provided as color design sheets. 
The training set consists of 12 characters and 11,345 frames, while the test set contains 10 characters and 3,000 frames. 
Image resolutions are $1920\times1920$ for training and $1024\times1024$ for testing.
The hand-drawn test data, sourced from professional animations and paint software tutorials, includes 20 clips featuring different characters with a total of 200 frames. 
The image sizes of hand-drawn test data range from $512\times512$ to $1600\times1600$.

To evaluate the proposed method, we employ segment-level and pixel-level accuracy. 
Reported values represent the average accuracy across all frames. 
In anime production, since pixels within individual segments can be collectively colored using software, segment-level accuracy reflects the cost of correcting miscoloring, while pixel-level accuracy measures visual quality.

\subsection{Implementation Details}
We use a single DACoN model for both keyframe and consecutive frame colorization tasks (\ie, the model weights are identical).
The reference images are randomly sampled from frames within the range of $\pm 0$ to $\pm 2$ relative to the target image in each epoch.
To enhance memory efficiency, images and segment masks are resized to $512 \times 512$ during training, and DINO features are pre-extracted. No resizing is applied during testing.
The model is trained for 5 epochs with a batch size of 2. We employ Adam optimizer~\cite{kingma2014adam} with a learning rate of $1\times10^{-4}$.
We use the officially released pre-trained DINOv2 Large model (\texttt{dinov24\_vitl14})~\cite{oquab2023dinov2}. 
The U-Net consists of 4 encoder and 4 decoder layers.
All experiments are conducted on a single NVIDIA GeForce RTX 4090 GPU, with a total training time of 14 hours.
Inference speed and memory usage details are provided in the supplementary material.

\begin{table}[t]
\centering
\caption{Quantitative comparison of keyframe colorization. \dq{Ref. shot} is the number of reference images, and \dq{max} corresponds to the case where all images in the color design sheet for each character are used. \dq{Acc-Thresh} is the accuracy computed only for segments larger than 10 pixels. \dq{Pix-Acc}, \dq{Pix-F-Acc}, and \dq{Pix-B-MIoU} represent pixel-wise accuracy, foreground pixel-wise accuracy, and pixel-wise background MIoU, respectively.}
\label{tab:keyframe_compare}
\resizebox{0.5\textwidth}{!}{
\begin{tabular}{lcccccccll}
\cline{1-8}
Method       & Ref. shot &                      & Acc                        & Acc-Thresh     & Pix-Acc        & Pix-F-Acc      & Pix-B-MIoU     &  &  \\ \cline{1-2} \cline{4-8}
ColorFlow    & 1         &                      & 12.10                      & 13.13          & 53.75          & 7.51           & 58.22          &  &  \\
MangaNinja   & 1         &                      & 17.39                      & 19.52          & 8.68           & 34.51          & 0.01           &  &  \\
AniDoc       & 1         &                      & 23.98                      & 27.06          & 78.71          & 50.25          & 86.64          &  &  \\
BasicPBC     & 1         &                      & \textemdash & 54.30          & 90.12          & 66.77          & 97.28          &  &  \\
BasicPBC-Ref & 1         &                      & \textemdash & 63.46          & 95.31          & 82.69          & 98.45          &  &  \\
Ours         & 1         &                      & \textbf{70.61}             & \textbf{75.15} & \textbf{97.82} & \textbf{92.16} & \textbf{99.66} &  &  \\ \cline{1-2} \cline{4-8}
ColorFlow    & 5         &                      & 13.19                      & 14.83          & 52.89          & 11.58          & 58.88          &  &  \\
BasicPBC-Ref & 5         &                      & \textemdash & 64.59          & 96.12          & 83.17          & 98.67          &  &  \\
Ours         & 5         & \multicolumn{1}{l}{} & \textbf{75.49}             & \textbf{79.55} & \textbf{98.43} & \textbf{94.72} & \textbf{99.60} &  &  \\ \cline{1-2} \cline{4-8}
Ours         & max       &                      & 76.66                      & 80.68          & 98.60          & 95.04          & 99.68          &  &  \\ \cline{1-8}
\end{tabular}
}
\end{table}

\begin{table}[t]
\centering
\caption{Quantitative results of keyframe colorization on the full test set.}
\label{tab:keyframe_compare_fulldata}
\resizebox{0.5\textwidth}{!}{
\begin{tabular}{lcccccccll}
\cline{1-8}
Method     & Ref. shot &                      & Acc            & Acc-Thresh     & Pix-Acc        & Pix-F-Acc      & Pix-B-MIoU     &  &  \\ \cline{1-2} \cline{4-8}
ColorFlow  & 1         &                      & 9.72           & 10.81          & 50.64          & 9.16           & 57.17          &  &  \\
MangaNinja & 1         &                      & 14.86          & 16.73          & 7.11           & 28.52          & 0.00           &  &  \\
AniDoc     & 1         &                      & 19.80          & 22.68          & 77.38          & 46.46          & 87.32          &  &  \\
Ours       & 1         &                      & \textbf{67.87} & \textbf{72.58} & \textbf{96.99} & \textbf{91.00} & \textbf{99.08} &  &  \\ \cline{1-2} \cline{4-8}
ColorFlow  & 5         &                      & 12.64          & 14.37          & 54.51          & 15.26          & 61.22          &  &  \\
Ours       & 5         & \multicolumn{1}{l}{} & \textbf{73.25} & \textbf{77.44} & \textbf{97.74} & \textbf{93.70} & \textbf{99.13} &  &  \\ \cline{1-2} \cline{4-8}
Ours       & max       &                      & 74.31          & 78.48          & 98.04          & 94.27          & 99.10          &  &  \\ \cline{1-8}
\end{tabular}
}
\end{table}

\begin{figure*}[t]
    \centering
    \begin{minipage}{\textwidth}
        \centering
        \includegraphics[width=1.0\linewidth]{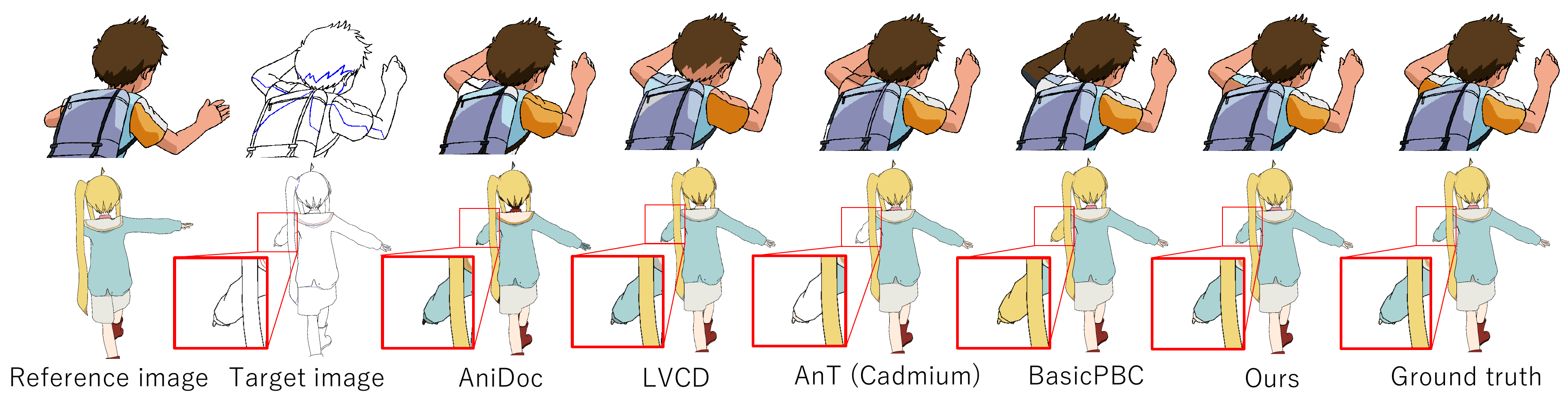}
        \caption{Visual comparisons of consecutive frame colorization between DACoN and other approaches. The top row shows colorized examples of hand-drawn line drawings, while the bottom row presents colorized examples of 3D rendered line drawings. DACoN accurately colors even small segments by effectively capturing semantic information.}
        \label{fig:consecutive_frame_compare}
    \end{minipage}
    \vspace{5mm}
    
    \begin{minipage}{\textwidth}
       \captionof{table}{Quantitative comparison for consecutive frame colorization. \dq{Mono} refers to the accuracy when color information is removed from the line drawings during testing.}
\centering
\label{tab:consecutive_frame_compare}
\resizebox{\textwidth}{!}{
\begin{tabular}{lcccccccccccc}
\hline
\multirow{2}{*}{Method}    &                      & \multicolumn{5}{c}{3D rendered}                                                    &           & \multicolumn{5}{c}{Hand-drawn}                                                     \\ \cline{3-7} \cline{9-13} 
                           &                      & Acc            & Acc-Thresh     & Pix-Acc        & Pix-F-Acc      & Pix-B-MIoU     &           & Acc            & Acc-Thresh     & Pix-Acc        & Pix-F-Acc      & Pix-B-MIoU     \\ \cline{1-1} \cline{3-7} \cline{9-13} 
AniDoc                     & \multicolumn{1}{l}{} & 31.20          & 36.36          & 92.01          & 70.68          & 96.03          &           & 40.37          & 43.34          & 91.25          & 74.66          & 88.47          \\
LVCD                       & \multicolumn{1}{l}{} & 37.63          & 43.77          & 66.20          & 80.95          & 58.75          &           & 43.30          & 46.38          & 71.09          & 79.74          & 59.92          \\ \cline{1-1} \cline{3-7} \cline{9-13} 
AnT (Cadmium)              &                      & 66.34          & 77.13          & 97.65          & 94.71          & 98.01          &           & 72.24          & 78.79          & 97.92          & 93.03          & 98.99          \\
BasicPBC                   &                      & 82.66          & 87.26          & 99.05          & 97.24          & 99.48          &           & 85.93          & 89.29          & 99.00          & 96.38          & 99.84          \\ \cline{1-1} \cline{3-7} \cline{9-13} 
Ours                       &                      & \textbf{84.76} & \textbf{88.23} & \textbf{99.27} & \textbf{97.97} & 99.63          & \textbf{} & \textbf{87.44} & \textbf{90.48} & \textbf{99.19} & \textbf{96.91} & 99.83          \\
Ours (Mono)                & \multicolumn{1}{l}{} & 83.59          & 87.26          & 99.23          & 97.86          & 99.63          &           & 86.09          & 89.16          & 99.09          & 96.54          & 99.85          \\
Ours w/o $L_{\mathrm{dc}}$ &                      & 84.06          & 87.73          & 99.22          & 97.74          & \textbf{99.64} &           & 86.95          & 90.10          & 99.02          & 96.65          & \textbf{99.87} \\ \hline
\end{tabular}
}
 
    \end{minipage}
\end{figure*}

\begin{figure*}[t]
    \centering
    \includegraphics[width=0.9\linewidth]{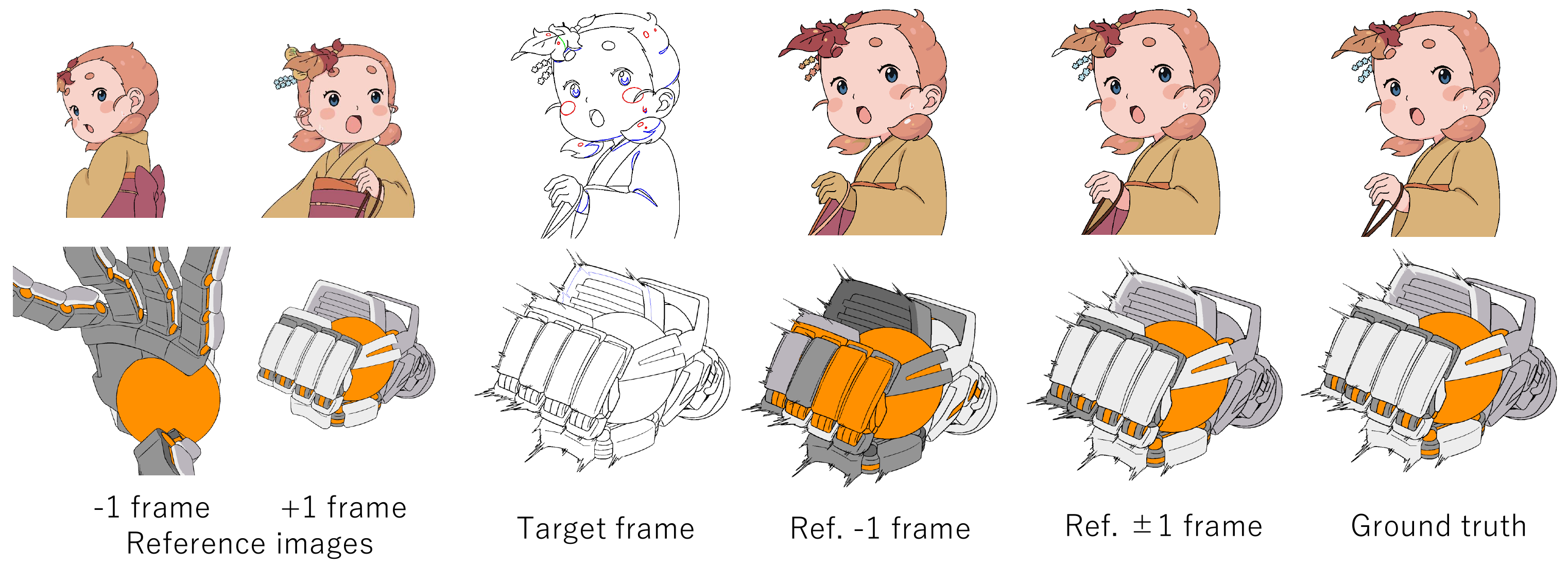}
    \caption{Visual comparison based on different reference frames for consecutive frame colorization.}
    \label{fig:consecutive_frame_reference_data}
\end{figure*}

\subsection{Keyframe Colorization}
Given that there has been limited work on keyframe colorization, we select baselines from two related categories. For segment-based methods, we include BasicPBC~\cite{dai2024learning} and BasicPBC-Ref~\cite{dai2024paint}, the latter being the only existing approach explicitly designed for keyframe colorization.
In addition, we choose three generation-based methods---ColorFlow~\cite{zhuang2024colorflow}, MangaNinja~\cite{liu2025manganinja}, and AniDoc~\cite{meng2024anidoc}---as baselines, given their prominence in colorization tasks.
In keyframe colorization evaluation, character color design sheets serve as reference images. Since hand-drawn data lacks these sheets, comparisons are conducted solely on 3D rendered data.

We use the reported results for both BasicPBC and BasicPBC-Ref from the BasicPBC-Ref paper~\cite{dai2024paint}.
Since BasicPBC-Ref was evaluated on only 8 of the 10 test characters due to its data requirements, Table~\ref{tab:keyframe_compare} shows comparisons on this restricted set, while Table~\ref{tab:keyframe_compare_fulldata} includes results over all test characters.\footnote{In the ICCV camera-ready version, only the results of BasicPBC and BasicPBC-Ref were reported based on a subset of the test set. This detail was not explicitly mentioned in the original papers~\cite{dai2024learning,dai2024paint}, and we found this after checking the released code. For fair comparison, we now provide two tables: Table~\ref{tab:keyframe_compare} reports results on the subset, and Table~\ref{tab:keyframe_compare_fulldata} reports results on the full test set (for consistency with the ICCV version).}
We conduct our own experiments on the generation-based baselines using their official pre-trained weights.
As image and video generation-based methods operate at the pixel level, we apply a post-processing step (Figure~\ref{fig:generate_method_post_process}) to enable segment-level comparison.
The detailed colorization process for each method is as follows. 
For ColorFlow, we use weights optimized for sketches and convert the target image to grayscale before input. 
For MangaNinja, we disable user-provided correspondence points and use only the grayscaled target image as input.
For AniDoc, as it is a video generation method that requires at least 14 conditioning frames, we provide a sequence of 14 binarized target images (all identical) and use the first generated frame for evaluation. 

Table~\ref{tab:keyframe_compare} presents quantitative comparisons, demonstrating that the proposed method outperforms the baselines.
Notably, our method achieves high accuracy (Acc), including for small segments of 10 pixels or less.
As shown in Figure~\ref{fig:keyframe_compare}, generation-based methods tend to over-color the background. 
Additionally, due to their reliance on training image resolution, they struggle to capture fine details in high-resolution images, such as eye highlights.
In contrast, our method preserves fine details such as shoelaces, even under significant viewpoint changes.

The effect of increasing reference images varies across methods.
Our method shows substantial improvement with five reference images, while the performance of BasicPBC-Ref remains largely unchanged.
This difference likely stems from BasicPBC-Ref selecting reference images similar to the target image before segment matching, whereas our method matches segments across all reference images, thereby leveraging additional information for colorization. 
Furthermore, as illustrated in Figure~\ref{fig:keyframe_reference_counts}, even the inclusion of less similar reference images does not degrade performance. This indicates that per-frame reference selection may be unnecessary, which simplifies the overall process.
Additional visual examples, including difficult cases and diverse poses, are available in the supplementary materials.

\subsection{Consecutive Frame Colorization}
We compare DACoN with previous segment correspondence based approaches designed for consecutive frame colorization~\cite{dai2024paint,casey2021animation}.
Although the training code for AnT~\cite{casey2021animation} is not publicly available, an official implementation is provided in the Cadmium application~\cite{Cadmium}, which we use for inference. Specifically, version 0.3.1 of Cadmium was used for quantitative comparisons, while version 0.3.4 was used for qualitative comparisons.
For generation-based methods (\ie, AniDoc~\cite{meng2024anidoc} and LVCD~\cite{huang2024lvcd}), we follow the same setup used in keyframe colorization by inputting 14 target images (identical) and using the first generated frame for evaluation.

Quantitative results for the AnT and BasicPBC methods are obtained from the BasicPBC paper~\cite{dai2024paint}, while qualitative comparisons are conducted using inference results from the officially distributed models.
For evaluation, we use the line drawing of the target image's $-1$st frame as the reference, and colorize all subsequent frames of each clip.

\begin{table}[t]
\centering
\caption{Quantitative comparison of consecutive frame colorization in our proposed method using different reference data. \dq{$-1$ frame} and \dq{$\pm 1$ frame} refer to the preceding frame and the preceding and following frames relative to the target image, respectively. \dq{1 shot} and \dq{5 shot} indicate the number of reference images from the color design sheet.}
\label{tab:consecutive_frame_reference_data}
\resizebox{0.5\textwidth}{!}{
\begin{tabular}{lccccccll}
\cline{1-7}
\multirow{2}{*}{Reference data} &                      & \multicolumn{5}{c}{3D rendered}                                                                                  &           &  \\ \cline{3-7}
                                &                      & Acc                  & Acc-Thresh           & Pix-Acc              & Pix-F-Acc            & Pix-B-MIoU           &           &  \\ \cline{1-1} \cline{3-7}
$-1$ frame                      &                      & 84.71                & 88.16                & 99.26                & 97.96                & 99.62                &           &  \\
$-1$ frame and 1 shot           &                      & 85.41                & 88.95                & 99.37                & 98.26                & \textbf{99.65}       &           &  \\
$-1$ frame and 5 shot           &                      & \textbf{85.61}       & \textbf{89.14}       & \textbf{99.38}       & \textbf{98.27}       & 99.64                &           &  \\ \cline{1-1} \cline{3-7}
$\pm 1$ frame                   &                      & \textbf{88.93}       & 91.91                & \textbf{99.62}       & 98.92                & \textbf{99.80}       & \textbf{} &  \\
$\pm 1$ frame and 1 shot        &                      & 88.91                & \textbf{91.92}       & \textbf{99.62}       & \textbf{98.93}       & \textbf{99.80}       &           &  \\
$\pm 1$ frame and 5 shot        &                      & 88.83                & 91.88                & \textbf{99.62}       & \textbf{98.93}       & 99.79                &           &  \\ \cline{1-7}
                                & \multicolumn{1}{l}{} & \multicolumn{1}{l}{} & \multicolumn{1}{l}{} & \multicolumn{1}{l}{} & \multicolumn{1}{l}{} & \multicolumn{1}{l}{} &           &  \\ \cline{1-7}
\multirow{2}{*}{Reference data} & \multicolumn{1}{l}{} & \multicolumn{5}{c}{Hand-drawn}                                                                                   &           &  \\ \cline{3-7}
                                & \multicolumn{1}{l}{} & Acc                  & Acc-Thres            & Pix-Acc              & Pix-F-Acc            & Pix-B-MIoU           &           &  \\ \cline{1-1} \cline{3-7}
$-1$ frame                      & \multicolumn{1}{l}{} & 87.02                & 90.10                & 99.15                & 96.75                & 99.83                &           &  \\
$\pm 1$ frame                   & \multicolumn{1}{l}{} & \textbf{91.32}       & \textbf{93.93}       & \textbf{99.65}       & \textbf{98.68}       & \textbf{99.93}       &           &  \\ \cline{1-7}
\end{tabular}
}
\vspace{-0.4cm}
\end{table}

The results are presented in Table~\ref{tab:consecutive_frame_compare} and Figure~\ref{fig:consecutive_frame_compare}.
Quantitative comparisons show an improvement in accuracy over the previous methods.
This improvement is attributed to our method’s ability to map occluded parts from the preceding reference frame to semantically corresponding parts, leading to better colorization.

As an additional experiment, we incorporate both the $+1$st frame and color design sheets into the reference images alongside the $-1$st frame. 
This experiment aims to test the method’s performance under ideal conditions, where all parts to be colorized in the target image are present in the reference images, and segment shapes and sizes are similar. 
In this setup, all frames except the first and last frames of each clip are used.

The results are presented in Table~\ref{tab:consecutive_frame_reference_data}.
The findings indicate that referring to both the preceding and following frames significantly improves accuracy compared to using only the preceding frame, for both 3D rendered and hand-drawn data. 
As shown in Figure~\ref{fig:consecutive_frame_reference_data}, in scenes where the character turns around or undergoes significant deformation, parts missing in the preceding frame appear in the following frame. 
This comprehensive reference information, combined with the method's ability to establish appropriate correspondences, likely drives the improvement in accuracy.
Regarding the color design sheet, adding it to the preceding frame leads to a slight improvement in accuracy. However, adding it to both the preceding and following frames, where all necessary part information is already available, showed no significant change in accuracy. 
This suggests that, similar to keyframe colorization, our method effectively handles reference images even when they may introduce unnecessary noise for colorization.

\begin{figure}[t]
    \centering
    \includegraphics[width=1.0\linewidth]{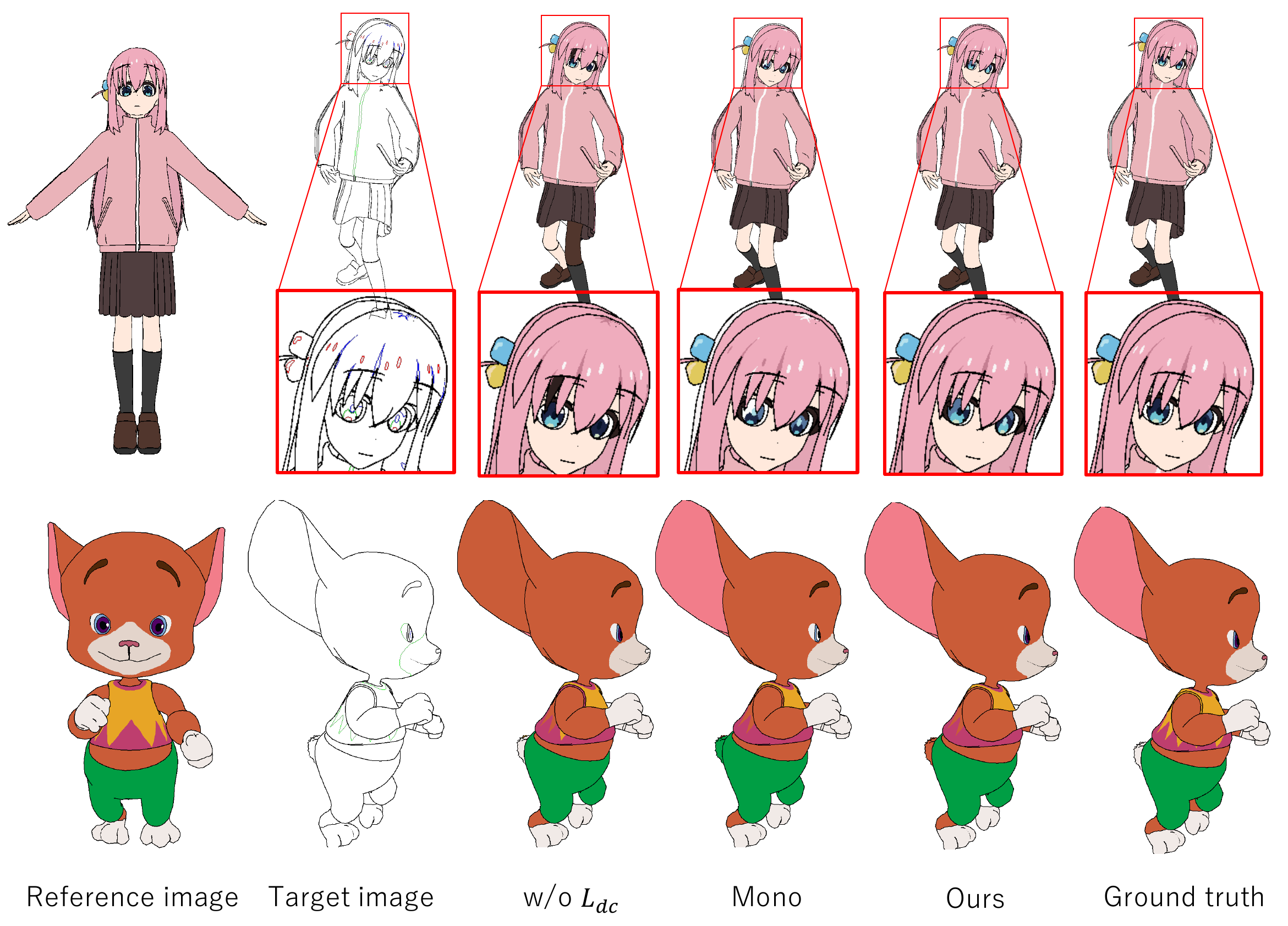}
    \caption{Ablation study of different components in our method for keyframe colorization.}
    \label{fig:keyframe_ablation}
\end{figure}
\begin{table}[t]
\centering
\caption{Quantitative ablation study of keyframe colorization.}
\label{tab:keyframe_ablation}
\resizebox{0.5\textwidth}{!}{
\begin{tabular}{lcccccccll}
\cline{1-8}
Method                     & Ref. shot &                      & Acc            & Acc-Thresh     & Pix-Acc        & Pix-F-Acc      & Pix-B-MIoU     &  &  \\ \cline{1-2} \cline{4-8}
Ours                       & 1         &                      & \textbf{67.87} & \textbf{72.58} & \textbf{96.99} & \textbf{91.00} & 99.08          &  &  \\
Ours (Mono)                & 1         &                      & 64.32          & 69.16          & 96.73          & 90.25          & 99.10          &  &  \\
Ours w/o $L_{\mathrm{dc}}$ & 1         & \multicolumn{1}{l}{} & 63.95          & 68.52          & 96.04          & 88.01          & \textbf{99.19} &  &  \\ \cline{1-2} \cline{4-8}
Ours                       & 5         & \multicolumn{1}{l}{} & \textbf{73.25} & \textbf{77.44} & \textbf{97.74} & \textbf{93.70} & 99.13          &  &  \\
Ours (Mono)                & 5         & \multicolumn{1}{l}{} & 70.26          & 74.70          & 97.62          & 93.37          & 99.06          &  &  \\
Ours w/o $L_{\mathrm{dc}}$ & 5         &                      & 72.57          & 76.77          & 97.83          & 93.42          & \textbf{99.22} &  &  \\ \cline{1-8}
\end{tabular}
}
\vspace{-0.2cm}
\end{table}

\subsection{Ablation Study}
To evaluate the effectiveness of the DINO-guided Feature Consistency Loss, we compare models trained with and without this loss. 
Additionally, we assess the impact of color information in the input line drawings, which use different line colors to represent highlights and shadows similar to those in anime production. To do this, we also compare the accuracy when all lines are unified to black during inference (Ours (Mono)). 

The quantitative results (see Tables~\ref{tab:consecutive_frame_compare} and~\ref{tab:keyframe_ablation}) demonstrate that the DINO-guided Feature Consistency Loss significantly improves accuracy in both keyframe and consecutive frame colorization, with a particularly notable improvement in keyframe colorization.
This improvement can be attributed to the ability of DINOv2 to better preserve semantic features when guided by the consistency loss.
Furthermore, as illustrated in Figure~\ref{fig:keyframe_ablation}, when color information is removed from the line drawings, shading and highlights in critical areas such as the hair and eyes become less distinct. This highlights the importance of color information for distinguishing segments with similar shapes and positions.

\subsection{Limitation}
\begin{figure}[t]
    \centering
    \includegraphics[width=1.0\linewidth]{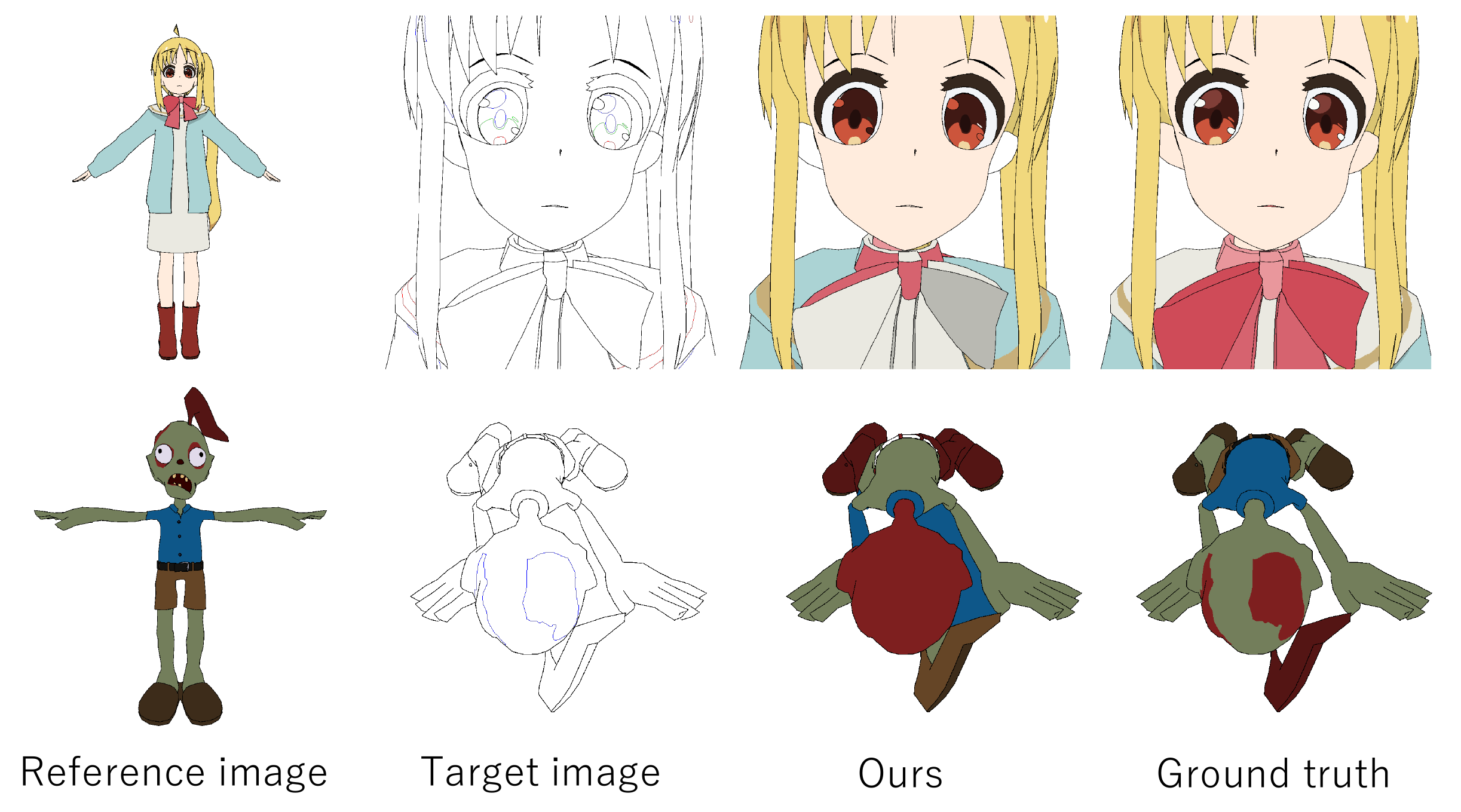}
    \caption{Failure cases for keyframe colorization.}
    \vspace{-0.2cm}
    \label{fig:limitation}
\end{figure}

Our method performs well in keyframe colorization but still faces several challenges.
First, when characters appear close to the camera, some body parts may fall outside the frame, preventing the U-Net from accessing complete line drawings (Figure~\ref{fig:limitation}, top).
Second, extreme poses can reduce accuracy since the color design sheets mainly depict upright stances, limiting useful reference information (Figure~\ref{fig:limitation}, bottom).
Finally, small background regions surrounded by multiple foreground parts can cause foreground-background misidentifications, resulting in incorrect colorization.
These limitations highlight areas for future work, such as improving the handling of off-screen features and extreme poses, and refining segmentation to better distinguish between foreground and background.

\section{Conclusion}
In this paper, we propose an automatic colorization method for anime line drawings that utilizes feature representations from a foundation model to guide segment correspondence across multiple reference images.
Our method outperforms previous approaches in both keyframe and consecutive frame colorization, using a unified model and training process. 
Furthermore, incorporating more reference images allows access to more comprehensive part information, leading to significant improvements in both tasks.

We hope this work highlights the potential of foundation models to extract effective features from line drawings, and encourages further research on tasks that support creative activities.

{
    \small
    \bibliographystyle{ieeenat_fullname}
    \bibliography{main}

\begin{thebibliography}{48}
\providecommand{\natexlab}[1]{#1}
\providecommand{\url}[1]{\texttt{#1}}
\expandafter\ifx\csname urlstyle\endcsname\relax
  \providecommand{\doi}[1]{doi: #1}\else
  \providecommand{\doi}{doi: \begingroup \urlstyle{rm}\Url}\fi

\bibitem[Cad()]{Cadmium}
Cadmium.
\newblock \url{https://cadmium.app/}.

\bibitem[Cao et~al.(2021)Cao, Mo, and Gao]{ruizhi2021line}
Ruizhi Cao, Haoran Mo, and Chengying Gao.
\newblock Line art colorization based on explicit region segmentation.
\newblock \emph{Computer Graphics Forum}, 40\penalty0 (7):\penalty0 1--10, 2021.

\bibitem[Cao et~al.(2024)Cao, Meng, Mok, Lee, Liu, and Li]{cao2024animediffusion}
Yu Cao, Xiangqiao Meng, PY Mok, Tong-Yee Lee, Xueting Liu, and Ping Li.
\newblock {AnimeDiffusion}: Anime diffusion colorization.
\newblock \emph{IEEE Transactions on Visualization and Computer Graphics}, 30\penalty0 (10):\penalty0 6956--6969, 2024.

\bibitem[Casey et~al.(2021)Casey, P{\'e}rez, and Li]{casey2021animation}
Evan Casey, V{\'\i}ctor P{\'e}rez, and Zhuoru Li.
\newblock The animation transformer: Visual correspondence via segment matching.
\newblock In \emph{Proceedings of the IEEE/CVF International Conference on Computer Vision (ICCV)}, pages 11323--11332, 2021.

\bibitem[Chen et~al.(2020)Chen, Zhang, Gao, He, Xia, Shi, and Zhang]{chen2020active}
Shu-Yu Chen, Jia-Qi Zhang, Lin Gao, Yue He, Shihong Xia, Min Shi, and Fang-Lue Zhang.
\newblock Active colorization for cartoon line drawings.
\newblock \emph{IEEE Transactions on Visualization and Computer Graphics}, 28\penalty0 (2):\penalty0 1198--1208, 2020.

\bibitem[Dai et~al.(2024{\natexlab{a}})Dai, Li, Zhou, Luo, Li, and Loy]{dai2024paint}
Yuekun Dai, Qinyue Li, Shangchen Zhou, Yihang Luo, Chongyi Li, and Chen~Change Loy.
\newblock Paint bucket colorization using anime character color design sheets.
\newblock \emph{arXiv preprint arXiv:2410.19424}, 2024{\natexlab{a}}.

\bibitem[Dai et~al.(2024{\natexlab{b}})Dai, Zhou, Li, Li, and Loy]{dai2024learning}
Yuekun Dai, Shangchen Zhou, Qinyue Li, Chongyi Li, and Chen~Change Loy.
\newblock Learning inclusion matching for animation paint bucket colorization.
\newblock In \emph{Proceedings of the IEEE/CVF Conference on Computer Vision and Pattern Recognition (CVPR)}, pages 25544--25553, 2024{\natexlab{b}}.

\bibitem[Dang et~al.(2020)Dang, Do, Nguyen, Pham, Nguyen, Hoang, and Nguyen]{dang2020correspondence}
Trung~DQ Dang, Thien Do, Anh Nguyen, Van Pham, Quoc Nguyen, Bach Hoang, and Giao Nguyen.
\newblock Correspondence neural network for line art colorization.
\newblock In \emph{ACM SIGGRAPH 2020 Posters}, pages 1--2, 2020.

\bibitem[Furusawa et~al.(2017)Furusawa, Hiroshiba, Ogaki, and Odagiri]{furusawa2017comicolorization}
Chie Furusawa, Kazuyuki Hiroshiba, Keisuke Ogaki, and Yuri Odagiri.
\newblock Comicolorization: Semi-automatic manga colorization.
\newblock In \emph{SIGGRAPH Asia 2017 Technical Briefs}, pages 1--4, 2017.

\bibitem[Huang et~al.(2024)Huang, Zhang, and Liao]{huang2024lvcd}
Zhitong Huang, Mohan Zhang, and Jing Liao.
\newblock {LVCD}: Reference-based lineart video colorization with diffusion models.
\newblock \emph{ACM Transactions on Graphics (TOG)}, 43\penalty0 (6):\penalty0 1--11, 2024.

\bibitem[Kingma and Ba(2014)]{kingma2014adam}
Diederik~P Kingma and Jimmy Ba.
\newblock Adam: A method for stochastic optimization.
\newblock \emph{arXiv preprint arXiv:1412.6980}, 2014.

\bibitem[Lee et~al.(2020)Lee, Kim, Lee, Kim, Chang, and Choo]{lee2020reference}
Junsoo Lee, Eungyeup Kim, Yunsung Lee, Dongjun Kim, Jaehyuk Chang, and Jaegul Choo.
\newblock Reference-based sketch image colorization using augmented-self reference and dense semantic correspondence.
\newblock In \emph{Proceedings of the IEEE/CVF Conference on Computer Vision and Pattern Recognition (CVPR)}, pages 5801--5810, 2020.

\bibitem[Li et~al.(2022)Li, Geng, Kang, Chen, and Yang]{li2022eliminating}
Zekun Li, Zhengyang Geng, Zhao Kang, Wenyu Chen, and Yibo Yang.
\newblock Eliminating gradient conflict in reference-based line-art colorization.
\newblock In \emph{European conference on computer vision}, pages 579--596. Springer, 2022.

\bibitem[Lin et~al.(2024)Lin, Zhao, and Wang]{lin2024visual}
Jianxin Lin, Wei Zhao, and Yijun Wang.
\newblock Visual correspondence learning and spatially attentive synthesis via transformer for exemplar-based anime line art colorization.
\newblock \emph{IEEE Transactions on Multimedia}, 26:\penalty0 6880--6890, 2024.

\bibitem[Liu et~al.(2020)Liu, Wang, Wu, and Seah]{liu2020shape}
Shaolong Liu, Xingce Wang, Zhongke Wu, and Hock~Soon Seah.
\newblock Shape correspondence based on {Kendall} shape space and {RAG} for {2D} animation.
\newblock \emph{The Visual Computer}, 36:\penalty0 2457--2469, 2020.

\bibitem[Liu et~al.(2025)Liu, Cheng, Chen, Xiao, Ouyang, Zhu, Liu, Shen, Chen, and Luo]{liu2025manganinja}
Zhiheng Liu, Ka~Leong Cheng, Xi Chen, Jie Xiao, Hao Ouyang, Kai Zhu, Yu Liu, Yujun Shen, Qifeng Chen, and Ping Luo.
\newblock Manganinja: Line art colorization with precise reference following.
\newblock \emph{arXiv preprint arXiv:2501.08332}, 2025.

\bibitem[Loftsdottir and Guzdial(2022)]{loftsdottir2022sketchbetween}
Dagmar Loftsdottir and Matthew Guzdial.
\newblock Sketchbetween: Video-to-video synthesis for sprite animation via sketches.
\newblock In \emph{Proceedings of the 17th International Conference on the Foundations of Digital Games}, pages 1--7, 2022.

\bibitem[Maejima et~al.(2019)Maejima, Kubo, Funatomi, Yotsukura, Nakamura, and Mukaigawa]{maejima2019graph}
Akinobu Maejima, Hiroyuki Kubo, Takuya Funatomi, Tatsuo Yotsukura, Satoshi Nakamura, and Yasuhiro Mukaigawa.
\newblock Graph matching based anime colorization with multiple references.
\newblock In \emph{ACM SIGGRAPH 2019 Posters}, pages 1--2, 2019.

\bibitem[Maejima et~al.(2021)Maejima, Kubo, Shinagawa, Funatomi, Yotsukura, Nakamura, and Mukaigawa]{maejima2021anime}
Akinobu Maejima, Hiroyuki Kubo, Seitaro Shinagawa, Takuya Funatomi, Tatsuo Yotsukura, Satoshi Nakamura, and Yasuhiro Mukaigawa.
\newblock Anime character colorization using few-shot learning.
\newblock In \emph{SIGGRAPH Asia 2021 Technical Communications}, pages 1--4, 2021.

\bibitem[Maejima et~al.(2024)Maejima, Shinagawa, Kubo, Funatomi, Yotsukura, Nakamura, and Mukaigawa]{maejima2024continual}
Akinobu Maejima, Seitaro Shinagawa, Hiroyuki Kubo, Takuya Funatomi, Tatsuo Yotsukura, Satoshi Nakamura, and Yasuhiro Mukaigawa.
\newblock Continual few-shot patch-based learning for anime-style colorization.
\newblock \emph{Computational Visual Media}, 10\penalty0 (4):\penalty0 705--723, 2024.

\bibitem[Meng et~al.(2024)Meng, Ouyang, Wang, Wang, Wang, Cheng, Liu, Shen, and Qu]{meng2024anidoc}
Yihao Meng, Hao Ouyang, Hanlin Wang, Qiuyu Wang, Wen Wang, Ka~Leong Cheng, Zhiheng Liu, Yujun Shen, and Huamin Qu.
\newblock {AniDoc}: Animation creation made easier.
\newblock \emph{arXiv preprint arXiv:2412.14173}, 2024.

\bibitem[Ning et~al.(2023)Ning, Muyao, Zhi, Zhihui, Zhiyong, Zhaoyan, Bin, and Haojie]{wang2023coloring}
Wang Ning, Niu Muyao, Dou Zhi, Wang Zhihui, Wang Zhiyong, Ming Zhaoyan, Liu Bin, and Li Haojie.
\newblock Coloring anime line art videos with transformation region enhancement network.
\newblock \emph{Pattern Recognition}, 141:\penalty0 109562, 2023.

\bibitem[Oquab et~al.(2023)Oquab, Darcet, Moutakanni, Vo, Szafraniec, Khalidov, Fernandez, Haziza, Massa, El-Nouby, et~al.]{oquab2023dinov2}
Maxime Oquab, Timoth{\'e}e Darcet, Th{\'e}o Moutakanni, Huy Vo, Marc Szafraniec, Vasil Khalidov, Pierre Fernandez, Daniel Haziza, Francisco Massa, Alaaeldin El-Nouby, et~al.
\newblock {DINOv2}: Learning robust visual features without supervision.
\newblock \emph{arXiv preprint arXiv:2304.07193}, 2023.

\bibitem[Radford et~al.(2021)Radford, Kim, Hallacy, Ramesh, Goh, Agarwal, Sastry, Askell, Mishkin, Clark, et~al.]{radford2021learning}
Alec Radford, Jong~Wook Kim, Chris Hallacy, Aditya Ramesh, Gabriel Goh, Sandhini Agarwal, Girish Sastry, Amanda Askell, Pamela Mishkin, Jack Clark, et~al.
\newblock Learning transferable visual models from natural language supervision.
\newblock In \emph{Proceedings of the International Conference on Machine Learning (ICML)}, pages 8748--8763. PMLR, 2021.

\bibitem[Ramassamy et~al.(2018)Ramassamy, Kubo, Funatomi, Ishii, Maejima, Nakamura, and Mukaigawa]{ramassamy2018pre}
Sophie Ramassamy, Hiroyuki Kubo, Takuya Funatomi, Daichi Ishii, Akinobu Maejima, Satoshi Nakamura, and Yasuhiro Mukaigawa.
\newblock Pre-and post-processes for automatic colorization using a fully convolutional network.
\newblock In \emph{SIGGRAPH Asia 2018 Posters}, pages 1--2, 2018.

\bibitem[Ranftl et~al.(2021)Ranftl, Bochkovskiy, and Koltun]{ranftl2021vision}
Ren{\'e} Ranftl, Alexey Bochkovskiy, and Vladlen Koltun.
\newblock Vision transformers for dense prediction.
\newblock In \emph{Proceedings of the IEEE/CVF international conference on computer vision}, pages 12179--12188, 2021.

\bibitem[Ravi et~al.(2024)Ravi, Gabeur, Hu, Hu, Ryali, Ma, Khedr, R{\"a}dle, Rolland, Gustafson, et~al.]{ravi2024sam}
Nikhila Ravi, Valentin Gabeur, Yuan-Ting Hu, Ronghang Hu, Chaitanya Ryali, Tengyu Ma, Haitham Khedr, Roman R{\"a}dle, Chloe Rolland, Laura Gustafson, et~al.
\newblock {SAM 2}: Segment anything in images and videos.
\newblock \emph{arXiv preprint arXiv:2408.00714}, 2024.

\bibitem[Rombach et~al.(2022)Rombach, Blattmann, Lorenz, Esser, and Ommer]{rombach2022high}
Robin Rombach, Andreas Blattmann, Dominik Lorenz, Patrick Esser, and Bj{\"o}rn Ommer.
\newblock High-resolution image synthesis with latent diffusion models.
\newblock In \emph{Proceedings of the IEEE/CVF Conference on Computer Vision and Pattern Recognition (CVPR)}, pages 10684--10695, 2022.

\bibitem[Ronneberger et~al.(2015)Ronneberger, Fischer, and Brox]{ronneberger2015u}
Olaf Ronneberger, Philipp Fischer, and Thomas Brox.
\newblock {U-Net}: Convolutional networks for biomedical image segmentation.
\newblock In \emph{Proceedings of the International Conference on Medical Image Computing and Computer Assisted Intervention (MICCAI)}, pages 234--241. Springer, 2015.

\bibitem[Sarlin et~al.(2020)Sarlin, DeTone, Malisiewicz, and Rabinovich]{sarlin2020superglue}
Paul-Edouard Sarlin, Daniel DeTone, Tomasz Malisiewicz, and Andrew Rabinovich.
\newblock Superglue: Learning feature matching with graph neural networks.
\newblock In \emph{Proceedings of the IEEE/CVF conference on computer vision and pattern recognition}, pages 4938--4947, 2020.

\bibitem[Sato et~al.(2014)Sato, Matsui, Yamasaki, and Aizawa]{sato2014reference}
Kazuhiro Sato, Yusuke Matsui, Toshihiko Yamasaki, and Kiyoharu Aizawa.
\newblock Reference-based manga colorization by graph correspondence using quadratic programming.
\newblock In \emph{SIGGRAPH Asia 2014 Technical Briefs}, pages 1--4, 2014.

\bibitem[Schuurmans et~al.(2018)Schuurmans, Berman, and Blaschko]{schuurmans2018efficient}
Mathijs Schuurmans, Maxim Berman, and Matthew~B Blaschko.
\newblock Efficient semantic image segmentation with superpixel pooling.
\newblock \emph{arXiv preprint arXiv:1806.02705}, 2018.

\bibitem[Shaolong et~al.(2023)Shaolong, Xingce, Xiangyuan, Zhongke, and Hock]{shaolong2023shape}
Liu Shaolong, Wang Xingce, Liu Xiangyuan, Wu Zhongke, and Seah Hock, Soon.
\newblock Shape correspondence for cel animation based on a shape association graph and spectral matching.
\newblock \emph{Computational Visual Media}, 9\penalty0 (3):\penalty0 633--656, 2023.

\bibitem[Shi et~al.(2020)Shi, Zhang, Chen, Gao, Lai, and Zhang]{shi2020deep}
Min Shi, Jia-Qi Zhang, Shu-Yu Chen, Lin Gao, Yu-Kun Lai, and Fang-Lue Zhang.
\newblock Deep line art video colorization with a few references.
\newblock \emph{arXiv preprint arXiv:2003.10685}, 2020.

\bibitem[Shi et~al.(2022)Shi, Zhang, Chen, Gao, Lai, and Zhang]{shi2022reference}
Min Shi, Jia-Qi Zhang, Shu-Yu Chen, Lin Gao, Yu-Kun Lai, and Fang-Lue Zhang.
\newblock Reference-based deep line art video colorization.
\newblock \emph{IEEE Transactions on Visualization and Computer Graphics}, 29\penalty0 (6):\penalty0 2965--2979, 2022.

\bibitem[Shugrina et~al.(2019)Shugrina, Liang, Kar, Li, Singh, Singh, and Fidler]{shugrina2019creative}
Maria Shugrina, Ziheng Liang, Amlan Kar, Jiaman Li, Angad Singh, Karan Singh, and Sanja Fidler.
\newblock {Creative Flow+} dataset.
\newblock In \emph{Proceedings of the IEEE/CVF Conference on Computer Vision and Pattern Recognition (CVPR)}, pages 5384--5393, 2019.

\bibitem[Siyao et~al.(2021)Siyao, Zhao, Yu, Sun, Metaxas, Loy, and Liu]{siyao2021deep}
Li Siyao, Shiyu Zhao, Weijiang Yu, Wenxiu Sun, Dimitris Metaxas, Chen~Change Loy, and Ziwei Liu.
\newblock Deep animation video interpolation in the wild.
\newblock In \emph{Proceedings of the IEEE/CVF Conference on Computer Vision and Pattern Recognition (CVPR)}, pages 6587--6595, 2021.

\bibitem[Siyao et~al.(2022)Siyao, Li, Li, Dong, Liu, and Loy]{siyao2022animerun}
Li Siyao, Yuhang Li, Bo Li, Chao Dong, Ziwei Liu, and Chen~Change Loy.
\newblock {AnimeRun}: {2D} animation visual correspondence from open source {3D} movies.
\newblock \emph{Advances in Neural Information Processing Systems (NeurIPS)}, 35:\penalty0 18996--19007, 2022.

\bibitem[Tang et~al.(2023)Tang, Jia, Wang, Phoo, and Hariharan]{tang2023emergent}
Luming Tang, Menglin Jia, Qianqian Wang, Cheng~Perng Phoo, and Bharath Hariharan.
\newblock Emergent correspondence from image diffusion.
\newblock \emph{Advances in Neural Information Processing Systems (NeurIPS)}, 36:\penalty0 1363--1389, 2023.

\bibitem[Tang et~al.(2025)Tang, Guo, Liu, Wang, Hua, Zhong, Xiao, Huang, Song, Liang, et~al.]{tang2025generative}
Yunlong Tang, Junjia Guo, Pinxin Liu, Zhiyuan Wang, Hang Hua, Jia-Xing Zhong, Yunzhong Xiao, Chao Huang, Luchuan Song, Susan Liang, et~al.
\newblock Generative ai for cel-animation: A survey.
\newblock \emph{arXiv preprint arXiv:2501.06250}, 2025.

\bibitem[Wu et~al.(2023)Wu, Yan, Liu, Xu, and Zhang]{wu2023self}
Shukai Wu, Xiao Yan, Weiming Liu, Shuchang Xu, and Sanyuan Zhang.
\newblock Self-driven dual-path learning for reference-based line art colorization under limited data.
\newblock \emph{IEEE Transactions on Circuits and Systems for Video Technology}, 34\penalty0 (3):\penalty0 1388--1402, 2023.

\bibitem[Xing et~al.(2024)Xing, Liu, Xia, Zhang, Wang, Shan, and Wong]{xing2024tooncrafter}
Jinbo Xing, Hanyuan Liu, Menghan Xia, Yong Zhang, Xintao Wang, Ying Shan, and Tien-Tsin Wong.
\newblock {ToonCrafter}: Generative cartoon interpolation.
\newblock \emph{ACM Transactions on Graphics (TOG)}, 43\penalty0 (6):\penalty0 1--11, 2024.

\bibitem[Yang et~al.(2025)Yang, Fan, Lin, Wang, and Zhang]{yang2025layeranimate}
Yuxue Yang, Lue Fan, Zuzeng Lin, Feng Wang, and Zhaoxiang Zhang.
\newblock {LayerAnimate}: Layer-specific control for animation.
\newblock \emph{arXiv preprint arXiv:2501.08295}, 2025.

\bibitem[Yu et~al.(2024)Yu, Qian, Wang, Dong, and Liu]{yifeng2024animation}
Yifeng Yu, Jiangbo Qian, Chong Wang, Yihong Dong, and Baisong Liu.
\newblock Animation line art colorization based on the optical flow method.
\newblock \emph{Computer Animation and Virtual Worlds}, 35\penalty0 (1):\penalty0 2229, 2024.

\bibitem[Zhang et~al.(2024)Zhang, Herrmann, Hur, Polania~Cabrera, Jampani, Sun, and Yang]{zhang2023tale}
Junyi Zhang, Charles Herrmann, Junhwa Hur, Luisa Polania~Cabrera, Varun Jampani, Deqing Sun, and Ming-Hsuan Yang.
\newblock A tale of two features: {Stable Diffusion} complements {DINO} for zero-shot semantic correspondence.
\newblock \emph{Advances in Neural Information Processing Systems (NeurIPS)}, 36, 2024.

\bibitem[Zhang et~al.(2021)Zhang, Wang, Wen, Li, and Liu]{zhang2021line}
Qian Zhang, Bo Wang, Wei Wen, Hai Li, and Junhui Liu.
\newblock Line art correlation matching feature transfer network for automatic animation colorization.
\newblock In \emph{Proceedings of the IEEE/CVF Winter Conference on Applications of Computer Vision}, pages 3872--3881, 2021.

\bibitem[Zhu et~al.(2016)Zhu, Liu, Wong, and Heng]{zhu2016globally}
Haichao Zhu, Xueting Liu, Tien-Tsin Wong, and Pheng-Ann Heng.
\newblock Globally optimal toon tracking.
\newblock \emph{ACM Transactions on Graphics (TOG)}, 35\penalty0 (4):\penalty0 1--10, 2016.

\bibitem[Zhuang et~al.(2024)Zhuang, Ju, Zhang, Liu, Zhang, Yuan, and Shan]{zhuang2024colorflow}
Junhao Zhuang, Xuan Ju, Zhaoyang Zhang, Yong Liu, Shiyi Zhang, Chun Yuan, and Ying Shan.
\newblock {ColorFlow}: Retrieval-augmented image sequence colorization.
\newblock \emph{arXiv preprint arXiv:2412.11815}, 2024.

\end{thebibliography}
}
 
\newpage
\DisableBackref
\appendix

\setcounter{page}{1}

\newcommand{\myrebuttaltitle}{
   \begin{center}
      \Large \bf DACoN: DINO for Anime Paint Bucket Colorization\\ with Any Number of Reference Images \par
      {\Large \textnormal{Supplementary Material}}
      \vskip .375in
   \end{center}
}

\twocolumn[{%
\renewcommand\twocolumn[1][]{#1}%
\myrebuttaltitle

\begin{center}
\begin{minipage}{\textwidth}
    \centering
    \captionsetup{type=figure}
    \includegraphics[width=1.0\linewidth]{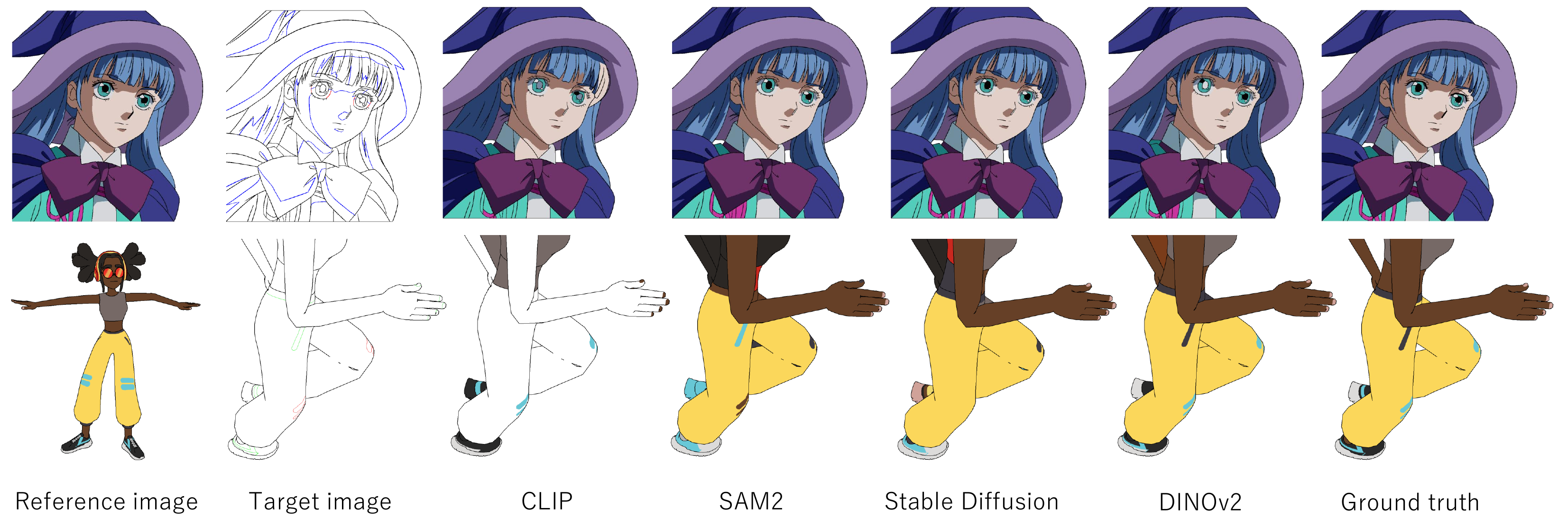}
    \captionof{figure}{Visual comparison of zero-shot colorization by foundational models. The top row shows the results of consecutive frame colorization, while the bottom row presents those of keyframe colorization.}
    \label{fig:foundation_models_compare}
\end{minipage}
\vspace{5mm}

\begin{minipage}{\textwidth}
   \centering
\captionof{table}{Quantitative comparison of zero-shot colorization by foundation models. \dq{Ours~(SD)} indicates the results obtained by replacing DINOv2 in our method with Stable Diffusion during training and evaluation.}
\label{tab:foundation_model_compare}
\resizebox{\textwidth}{!}{
\begin{tabular}{lcccccccccccc}
\hline
\multirow{2}{*}{Method}       &  & \multicolumn{5}{c}{Keyframe (3D rendered, 1-shot)}                                 &  & \multicolumn{5}{c}{Consecutive frame (Hand-drawn)}                                 \\ \cline{3-7} \cline{9-13} 
                              &  & Acc            & Acc-Thresh     & Pix-Acc        & Pix-F-Acc      & Pix-B-MIoU     &  & Acc            & Acc-Thresh     & Pix-Acc        & Pix-F-Acc      & Pix-B-MIoU     \\ \cline{1-1} \cline{3-7} \cline{9-13} 
CLIP                          &  & 36.72          & 38.95          & 86.76          & 60.46          & 88.37          &  & 67.13          & 69.56          & 97.07          & 88.65          & 99.00          \\
SAM2                          &  & 34.54          & 36.96          & 85.06          & 54.12          & 95.62          &  & 82.44          & 85.51          & 98.35          & 94.68          & 99.18          \\
Stable Diffusion              &  & 49.72          & 53.80          & 89.84          & 71.10          &  95.54              &  & \textbf{84.12} & \textbf{87.33} & 98.46          & \textbf{95.56} & 99.03          \\
DINOv2                        &  & \textbf{57.49} & \textbf{61.86} & \textbf{95.35} & \textbf{87.24} & \textbf{97.45} &  & 80.64          & 83.39          & \textbf{98.79} & 95.35          & \textbf{99.78} \\ \cline{1-1} \cline{3-7} \cline{9-13} 
Ours                          &  & 67.87          & 72.58          & 96.99          & 91.00          & 99.08          &  & 87.44          & 90.48          & 99.19          & 96.91          & 99.83          \\
Ours (SD)                     &  & 62.26          & 66.88          & 93.88          & 82.13          & 98.43          &  & 87.27          & 90.46          & 98.89          & 96.54          & 99.29          \\
Ours (SD) w/o $L_{\mathrm{dc}}$  &  & 60.96          & 65.52          & 93.43          & 80.57          & 98.79          &  & 86.93          & 90.23          & 98.64          & 95.99          & 99.74          \\ \hline
\end{tabular}
}
 
\end{minipage}
\vspace{5mm}
\end{center}
}]

\section{Comparisons with Foundation Models}
To evaluate the impact of DINOv2 features~\cite{oquab2023dinov2} on the proposed method, we assess zero-shot colorization performance using only DINO features---that is, without any additional training or fine-tuning.
Additionally, we compare DINOv2 with other visual foundation models to explore how their feature representations contribute to automatic colorization. 
For this comparison, we select Stable Diffusion~(SD)~\cite{rombach2022high}, which captures part-level semantic information similar to DINOv2~\cite{zhang2023tale}, as well as SAM2~\cite{ravi2024sam} and CLIP~\cite{radford2021learning}, which are widely used foundation models.

Segment pooling, as employed in our method, is applied to each model's feature map to enable segment correspondence and color propagation between the reference and target images.
The feature extraction methods for each model are as follows:
\begin{itemize}
    \item DINOv2: We use the Large model with an input size of $518 \times 518$, and features are extracted from the final encoder layer.
    \item SD: We use the Stable Diffusion v2-1 model with an input size of $768 \times 768$. The input text prompt is \dq{a photo of an anime character.} The feature map is extracted via the correspondence method proposed in~\cite{tang2023emergent} from the first layer of upsampling, with the dimensional step set to $261/1000$.
    \item SAM2: We use the Large model of SAM2.1 with an input size of $1024 \times 1024$, and features are extracted from the final encoder layer.
    \item CLIP: We select the ConvNext-Large model from OpenCLIP, as used in BasicPBC~\cite{dai2024paint}, with an input size of $224 \times 224$, and features are extracted from the final encoder layer.
\end{itemize}
In all models, both the reference and target images are line drawings.

The quantitative results are presented in Table~\ref{tab:foundation_model_compare}.
DINOv2 achieves the highest accuracy in keyframe colorization, while SD outperforms the others in consecutive frame colorization, suggesting that DINOv2 excels at capturing semantic, part-level information, whereas SD effectively leverages spatial details.
SAM2, as shown in Figure~\ref{fig:foundation_models_compare}, performs well in consecutive frame colorization, likely due to its training on high-resolution images, which helps capture fine-grained details.

Further experiments replacing DINOv2 with SD in the proposed method reveal that although SD performs well in zero-shot consecutive frame colorization, DINOv2 still achieves higher accuracy when integrated into our method.
This result indicates that the U-Net, which learns spatial features, complements DINOv2's strength in capturing semantic details.
In contrast, since both SD and the U-Net excel at capturing spatial information, their strengths overlap, thereby reducing the added benefit of using SD in this context.

In addition, ablation results show that our proposed Feature Consistency Loss improves performance not only with DINOv2 but also with other foundation models such as SD, which capture semantic information.

\begin{table}[t]
\centering
\caption{Comparison of inference time and memory usage during consecutive frame colorization on 3D synthetic test data.
\dq{Time} represents the average colorization time per sample, \dq{FPS} denotes the number of frames colorized per second, \dq{Params} indicates the model’s parameter size, and \dq{Peak Mem} refers to the peak memory usage during inference.
\dq{Ours*} corresponds to the configuration with segment pooling fixed at $512 \times 512$.
All experiments were conducted on an NVIDIA GeForce RTX 4090 GPU.}

\label{tab:time_and_memory_compare}
\resizebox{0.5\textwidth}{!}{
\begin{tabular}{llccccccll}
\cline{1-8}
Method          &  & Time        & FPS         & \multicolumn{1}{l}{} & Params  & Model Size & Peak Mem    &  &  \\
                &  & {[}ms{]}    &             & \multicolumn{1}{l}{} & {[}M{]} & {[}GB{]}   & {[}GB{]}    &  &  \\ \cline{1-1} \cline{3-4} \cline{6-8}
BasicPBC        &  & 1454.52     & 0.69        &                      & 26.33   & 0.10       & 2.62        &  &  \\
Ours            &  & 264.37      & 3.78        &                      & 339.85  & 1.30       & 6.01        &  &  \\
Ours*           &  & 249.11      & 4.01        &                      & 339.85  & 1.30       & 3.20        &  &  \\
Ours w/o DINOv2 &  & \textemdash & \textemdash &                      & 35.49   & 0.14       & \textemdash &  &  \\ \cline{1-8}
\end{tabular}
}
\end{table}
\begin{table}[t]
\centering
\caption{Quantitative comparison of different segment pooling sizes.
\dq{Ours} uses the same size as the input image, while \dq{Ours*} is fixed at $512 \times 512$.}

\label{tab:seg_pool_ablation}
\resizebox{0.5\textwidth}{!}{
\begin{tabular}{lccccccccll}
\cline{1-9}
Method &                      & \multicolumn{3}{c}{Keyframe (3D rendered, 1-shot)} &  & \multicolumn{3}{c}{Consecutive frame (Hand-drawn)} &  &  \\ \cline{3-5} \cline{7-9}
       & \multicolumn{1}{l}{} & Acc           & Acc-Thresh        & Pix-Acc        &  & Acc           & Acc-Thresh        & Pix-Acc        &  &  \\ \cline{1-1} \cline{3-5} \cline{7-9}
Ours   &                      & 67.87         & 72.58             & 96.99          &  & 87.44         & 90.48             & 99.19          &  &  \\
Ours*  &                      & 67.79         & 72.49             & 96.98          &  & 87.48         & 90.61             & 99.21          &  &  \\ \cline{1-9}
\end{tabular}
}
\end{table}
\section{Inference Time and Memory Usage}
To evaluate the implementation cost for anime production, we measure the inference time and memory usage of the proposed method and compare it with BasicPBC~\cite{dai2024learning}, which serves as our baseline.
A comparison with AnT~\cite{casey2021animation}, which also utilizes segment correspondence, is not possible because the internal processing details of the Cadmium application~\cite{Cadmium} are not publicly available.
For this comparison, consecutive frame colorization is performed on a test set of 2,850 samples of 3D rendered data. 
The measurements are taken from the moment the images are fed until the predicted color information for each segment is produced, with segment information pre-prepared and models preloaded into memory. 

The results are presented in Table~\ref{tab:time_and_memory_compare}.
Our proposed method is capable of colorizing approximately three samples per second, making it roughly five times faster than the baseline method.
This improvement can be attributed to the absence of optical flow estimation, which is required by previous approaches in addition to feature extraction.
However, since the segment region and color information are provided in advance for this dataset, the measured speed does not fully reflect the actual cost of automatic colorization.
In the case of test images where all segment regions are completely enclosed by line drawings, it takes an average of 1.67 seconds to extract the segment areas and corresponding colors.
As a result, the average time for complete automatic colorization is 3.12 seconds for BasicPBC and 1.9 seconds for our method.
According to prior research~\cite{maejima2021anime}, professional animators reported that automatic colorization of 20 to 30 frames within 5 to 10 minutes—accounting for manual correction time—is considered acceptable.
Therefore, both methods are practically viable in terms of speed.

On the other hand, in terms of memory consumption, our method relies on a large foundation model, resulting in a significantly larger model size than the baseline.
In particular, during segment pooling, the DINO feature map (with a feature dimension of 1024) is expanded to match the input image size, leading to a peak memory usage of over 6~GB during inference.
This suggests that memory usage may become a bottleneck when processing high-resolution target images.

To address this issue, we apply max pooling to downscale the segment masks to a resolution of $512 \times 512$, consistent with training conditions, and evaluate the performance under this setting (see \dq{Ours*} in Tables~\ref{tab:time_and_memory_compare} and~\ref{tab:seg_pool_ablation}).
As a result, we reduce the peak memory usage to approximately 3~GB---about half---without significantly sacrificing colorization accuracy.
This indicates that the proposed method can perform robust automatic colorization under memory constraints regardless of input resolution, further enhancing its practicality.

\section{Clip-Wise Colorization}
To further demonstrate the practical applicability of our method, we evaluate a scenario where only the first frame of each clip is provided as a reference.
As shown in Table~\ref{tab:first_frame_reference}, this setting significantly reduces accuracy compared to the consecutive frame colorization setting, where the previous frame (\ie, the $-1$st frame) is used as the reference.
Nevertheless, supplementing the first frame with additional color design sheets helps mitigate this gap, particularly improving accuracy in regions not visible in the first frame and highlights the strength of our multi-reference approach.
Improving performance under such limited-reference conditions remains a challenging problem, and leveraging temporal consistency across frames will be a key direction for future work.

\begin{table}[t]
\centering
\caption{Quantitative comparison of colorization using only the first frame of each clip or with additional reference images.}
\label{tab:first_frame_reference}
\resizebox{\linewidth}{!}{
\begin{tabular}{lllccccccc}
\hline
\multirow{2}{*}{Method} & \multirow{2}{*}{Ref. data} & \multicolumn{1}{c}{} & \multicolumn{3}{c}{3D rendered} & \multicolumn{1}{l}{} & \multicolumn{3}{c}{Hand-drawn}          \\ \cline{4-6} \cline{8-10} 
                        &                            & \multicolumn{1}{c}{} & Acc    & Acc-Thresh  & Pix-Acc  & \multicolumn{1}{l}{} & Acc         & Acc-Thresh  & Pix-Acc     \\ \cline{1-2} \cline{4-6} \cline{8-10} 
Ours                    & first frame                & \textbf{}            & 69.91  & 73.59       & 97.30    &                      & 65.85       & 69.22       & 93.50       \\ \cline{1-2} \cline{4-6} \cline{8-10} 
Ours                    & first frame + $1$ shot     &                      & 74.59  & 78.67       & 98.20    &                      & \textemdash & \textemdash & \textemdash \\
Ours                    & first frame + $5$ shot     &                      & 76.81  & 80.82       & 98.38    &                      & \textemdash & \textemdash & \textemdash \\
Ours                    & first frame + max shot     &                      & 77.28  & 81.28       & 98.39    &                      & \textemdash & \textemdash & \textemdash \\ \cline{1-2} \cline{4-6} \cline{8-10} 
Ours                    & $-1$ frame                 &                      & 84.66  & 88.23       & 99.27    &                      & 87.44       & 90.48       & 99.19       \\ \hline
\end{tabular}
}
\end{table}

\section{Ablation for Architectures}
We further investigate alternative architectures within our framework. Specifically, we evaluate the following three variants:

\begin{itemize}
    \item \textbf{DPT-based dimensionality reduction:} Following DPT~\cite{ranftl2021vision}, we aggregate features from multiple layers of DINOv2 (\ie, layers 5, 12, 18, and 24) and project them into a 128-dimensional space. The output resolution is set to $296 \times 296$.

    \item \textbf{Cross-attention-based feature fusion:} Instead of using simple concatenation followed by an MLP, we adopt a cross-attention mechanism to fuse CNN and DINO features. In this setup, CNN features serve as queries, and DINO features are used as keys and values. The module consists of 4 attention heads and 9 layers.

    \item \textbf{Multiplex Transformer integration:} We incorporate a Multiplex Transformer module---commonly used in prior work but deliberately not employed in our proposed method---to jointly process and aggregate segment features from both reference and target images for comparison purposes. This module also uses 4 attention heads and 9 layers.
\end{itemize}

\begin{table}[t]
\centering
\caption{Ablation results on architecture design.}
\label{tab:architecture_ablation}
\resizebox{\linewidth}{!}{
\begin{tabular}{llccclccc}
\hline
\multirow{2}{*}{Method}                  & \multicolumn{1}{c}{} & \multicolumn{3}{c}{Keyframe (3D rendered, 1-shot)} &                      & \multicolumn{3}{c}{Consecutive frame (Hand-drawn)} \\ \cline{3-5} \cline{7-9} 
                                         & \multicolumn{1}{c}{} & Acc             & Acc-Thresh      & Pix-Acc        &                      & Acc             & Acc-Thresh      & Pix-Acc        \\ \cline{1-1} \cline{3-5} \cline{7-9} 
Ours (MLP based)                         & \multicolumn{1}{c}{} & \textbf{67.87}  & \textbf{72.58}  & \textbf{96.99} & \textbf{}            & \textbf{87.44}  & \textbf{90.48}  & \textbf{99.19} \\ \cline{1-1} \cline{3-5} \cline{7-9} 
DPT (alt. dim.red.) &                      & 66.55           & 71.30           & 96.51          & \textbf{}            & 86.96           & 90.10           & 99.11          \\
Cross Atten. (alt. fusion)               &                      & 47.60           & 50.69           & 66.45          & \multicolumn{1}{c}{} & 85.59           & 88.79           & 98.41          \\
Multiplex Transformer (add. agg.)        &                      & 66.54           & 71.47           & 96.81          &                      & 87.08           & 89.99           & 99.09          \\ \hline
\end{tabular}
}
\end{table}

As summarized in Table~\ref{tab:architecture_ablation}, our MLP-based design consistently outperforms these more complex alternatives, including Transformer-based modules. 
Given that the evaluation dataset consists of 11,345 samples, with character diversity limited to 12 identities and the domain restricted to line drawings, we suggest that lightweight architectures with fewer parameters, such as MLPs, may offer a better trade-off under such constraints compared to heavier models like CNNs or Transformers.

\clearpage 
\begin{figure*}[t]
    \centering
    \includegraphics[width=1.0\linewidth]{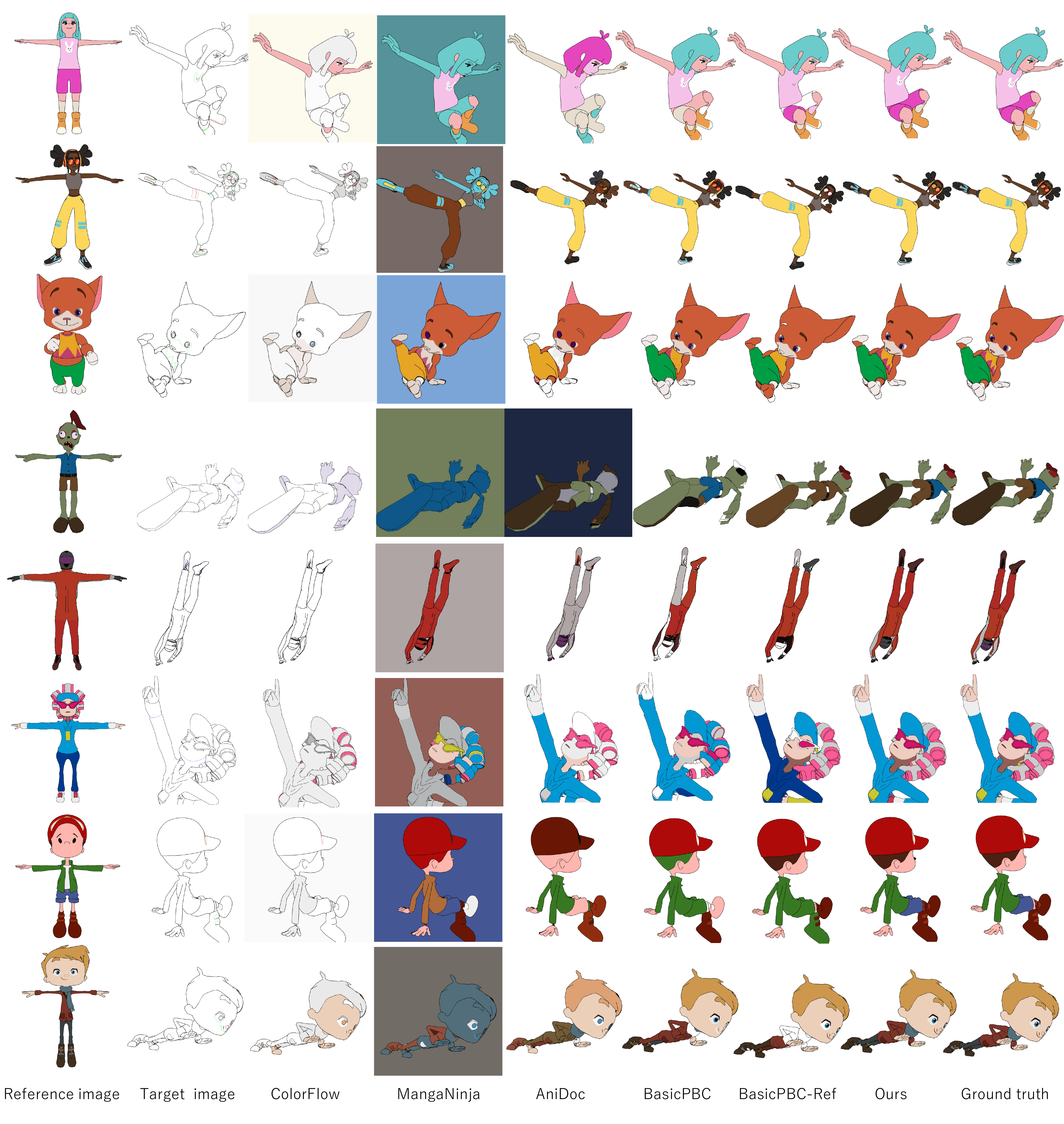}
    \caption{Additional qualitative comparisons of keyframe colorization results.}
    \label{fig:keyframe_compare_supply}
\end{figure*}
\clearpage

\end{document}